\documentclass[10pt,twocolumn,letterpaper]{article}

\usepackage[pagenumbers]{cvpr}      %
\usepackage[dvipsnames]{xcolor}
\usepackage{soul} %
\usepackage{lipsum}
\usepackage{xcolor}
\definecolor{lightblue}{rgb}{0.8,0.9,1}
\definecolor{lightgreen}{rgb}{0.9,1,0.9}
\definecolor{lightred}{rgb}{1,0.5,0.5}
\definecolor{gray}{rgb}{0.7,0.7,0.7}

\usepackage[T1]{fontenc}
\usepackage{newunicodechar}
\newunicodechar{—}{\textemdash}
\usepackage{graphicx}
\usepackage{flafter}
\usepackage{amsmath}
\usepackage{amssymb}  %
\usepackage{lipsum}
\usepackage{multirow}

\usepackage{amsthm}

\usepackage[ruled]{algorithm}
\usepackage{algpseudocode}

\usepackage{./rpm_packages/rpm_SIunits}
\usepackage{./rpm_packages/rpm_acronyms}
\usepackage{./rpm_packages/rpm_math}
\usepackage{./rpm_packages/rpm_misc}

\definecolor{skelblue}{rgb}{0.10,0.35,0.70}
\definecolor{slotbg}{rgb}{0.955,0.965,0.985}
\definecolor{slotgray}{rgb}{0.42,0.45,0.50}

\usepackage{placeins}
\renewcommand{\topfraction}{0.95}

\renewcommand{\textfraction}{0.05}
\usepackage{stfloats}

\definecolor{cvprblue}{rgb}{0.21,0.49,0.74}
\usepackage[pagebackref,breaklinks,colorlinks,citecolor=cvprblue]{hyperref}

\title{View Management with Gaussian Placement and Opacity Regularization for Online 3DGS}

\author{Dong Hwi Kim \quad Park Sangwon \quad Chaehyeon Song \quad Ayoung Kim\\
Seoul National University\\
{\tt\small \{martian35, swpants05, chaehyeon, ayoungk\}@snu.ac.kr}
}

\hypersetup{pdftitle={View Management with Gaussian Placement and Opacity Regularization for Online 3DGS}}

\begin{document}

\maketitle
\begin{abstract}
The realistic rendering of 3D Gaussian Splatting (3DGS) has drawn interest from real-time applications that require maps to be built online. This led to coupling 3DGS with SLAM, providing camera poses and initial Gaussians for the map. However, the map is still optimized with designs inherited from conventional SLAM, which train only on keyframes selected for localization and regularize geometry with uncertain frontend depth. These designs overlook the model complexity of a 3DGS map: with millions of parameters but few views of each region, the map overfits to sparse inputs. Resolving this is challenging, as every view and Gaussian added to the map competes for the same real-time budget. We close this gap from both sides. To add views while keyframes preserve the structure of the map, we progressively admit non-keyframe images as mapping updates complete and allocate the limited updates across different groups of frames according to the supervision each group provides. To control the model complexity of the map, edge-guided initialization places new Gaussians where image detail requires them. To complement the uncertain frontend depth, Self-Ref\-er\-enced Opac\-ity Reg\-u\-lar\-iza\-tion (SOR) uses accumulated rendering evidence, at no extra rendering cost, to suppress the opacity of Gaussians that persistently contribute in front of a map-derived surface reference. Together, these designs improve both the appearance and the geometry of online maps within the real-time budget.
\end{abstract}

\begin{figure}[t]
  \centering
  \includegraphics[width=\columnwidth]{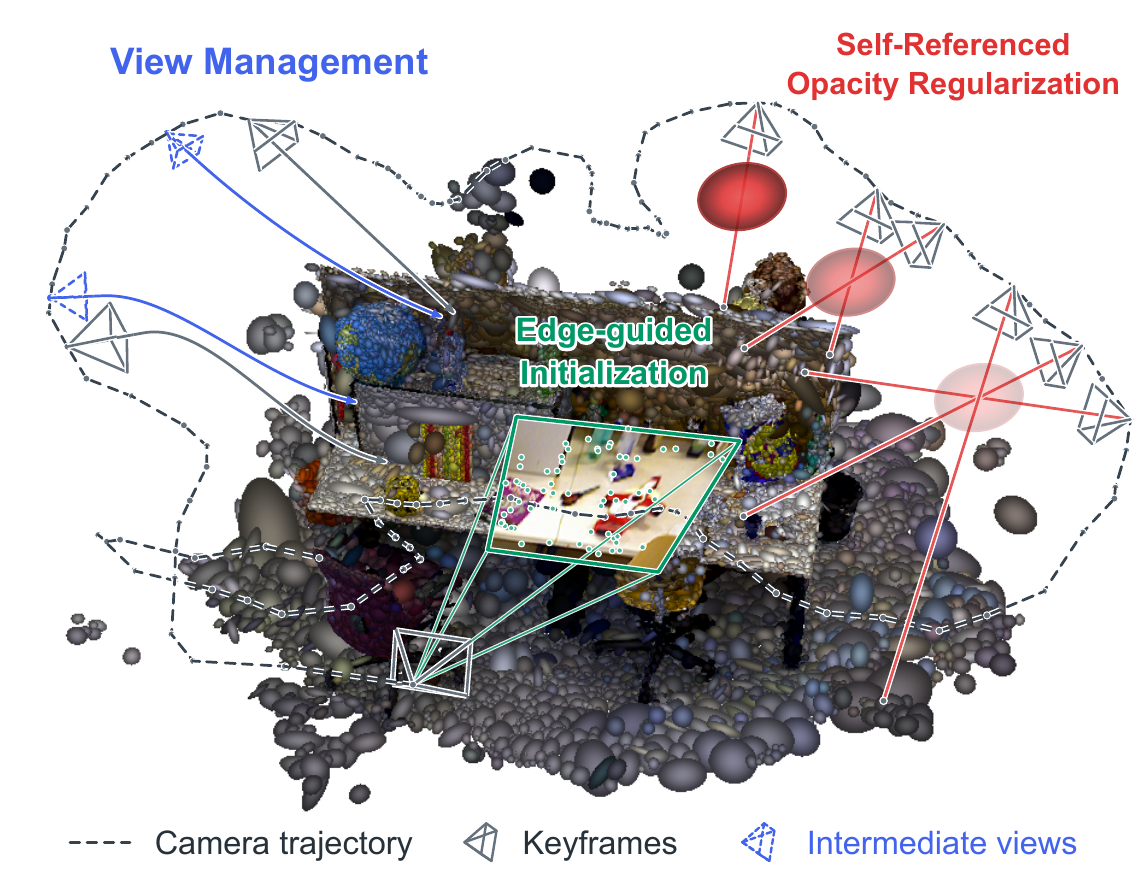}
  \caption{\textbf{Real-time Gaussian mapping.} Edge-guided initialization places Gaussians according to image detail, while view management refines the shared map from keyframes and intermediate views. Self-Referenced Opacity Regularization uses accumulated rendering evidence to suppress contributions in front of a map-derived surface, shown schematically by fading Gaussians.}
  \label{fig:scene_teaser}
\end{figure}

\section{Introduction}
\label{sec:intro}

Building geometrically accurate and photorealistic 3D maps online remains an important challenge in 3D vision and autonomous robotics.
Among the variety of 3D representations (including point clouds, meshes~\cite{lin2023immesh}, and surfels~\cite{chen2019suma}), 3D Gaussian Splatting (3DGS)~\cite{kerbl20233dgs} has gained attention for enabling dense, photorealistic maps with real-time rendering from novel viewpoints~\cite{matsuki2024monogs,huang2024photoslam}, offering opportunities for vision-based perception and decision-making~\cite{chen2025splat, intro_Shorinwa2025splatmover,intro_zitkovich2023rt2}.
To achieve this level of quality, the representation requires careful optimization of numerous parameters—including position, scale, rotation, color, and opacity—for millions of primitives.
Since this optimization process is computationally intensive and relies on dense multi-view constraints, 3DGS frameworks are predominantly trained offline using pre-collected datasets.
Autonomous systems, however, need to construct and update maps as they navigate and explore their environments, providing the spatial context beyond the current field of view to support timely planning and action.

Coupling 3DGS with SLAM has enabled online pose estimation and Gaussian initialization~\cite{matsuki2024monogs,huang2024photoslam,zhang2025hislam2,splatslam2024,vigsslam}: a frontend tracks the camera, selects keyframes, and estimates their depth, from which the mapper inserts Gaussians as keyframes arrive.
The Gaussian map itself, however, is still optimized with mapping designs inherited from conventional SLAM.
First, the map is optimized only on keyframes, which SLAM selects with little overlap to build a geometric map for localization.
Second, geometric regularization relies on frontend depth, which is uncertain where viewpoints are few.
These designs overlook the model complexity of a 3DGS map, which renders novel views and thus needs dense views with large overlap to constrain its parameters.
The map therefore overfits to sparse inputs, degrading at unseen viewpoints and retaining floaters, Gaussians that render in front of the true surface.
Offline, more views and longer training would resolve this; online, the mapper must keep pace with the stream, and each optimization step renders a single image, so every added view and every added Gaussian takes updates away from the rest of the map.
The problem is most pronounced in egocentric sequences, where head motion makes the observations of each region brief and intermittent, so a region often leaves the view before its Gaussians have converged.

Our method closes the gap between the model complexity of a 3DGS map and the views available to constrain it from both sides.
The dense views that a 3DGS map needs are already in the input stream: the images between consecutive keyframes largely overlap the keyframes.
We therefore progressively admit these non-keyframe images (blue in \cref{fig:scene_teaser}) into the training set as mapping updates complete, so that admitted images receive enough updates within the real-time budget.
Because the images provide different supervision but compete for the same steps, we allocate updates across three groups with fixed shares: recent keyframes refine newly initialized Gaussians, all keyframes revisit previously mapped regions as the frontend refines their poses and depth, and non-keyframe images constrain each region from additional viewpoints.
To control the model complexity of the map, edge-guided initialization places new Gaussians where image detail requires them (green in \cref{fig:scene_teaser}), since every Gaussian adds parameters for the available views to constrain and limited updates to refine.
To complement the uncertain frontend depth, Self-Referenced Opacity Regularization (SOR) accumulates evidence of where each surface lies from the views that training already renders, and suppresses the opacity of floaters, Gaussians that persistently contribute in front of this map-derived surface (red in \cref{fig:scene_teaser}).
In summary, we reduce overfitting in online 3DGS mapping within a real-time budget through the following contributions:
\begin{itemize}
    \item View management, which progressively admits the non-keyframe images between keyframes as mapping updates complete and allocates the limited updates across recent keyframes, all keyframes, and non-keyframe images according to the supervision each group provides.
    \item Edge-guided initialization, which uses image gradients to decide both how many Gaussians each keyframe adds relative to earlier keyframes and where they are placed within the image, so that the model complexity of the map follows the detail of the scene.
    \item Self-Referenced Opacity Regularization (SOR), which suppresses the opacity of Gaussians that persistently contribute in front of a surface reference accumulated from training renders, complementing uncertain frontend depth at no extra rendering cost.
    \item Experiments on egocentric and public sequences at their original frame rate, showing that our method improves both the appearance and the geometry of online maps over existing 3DGS-SLAM systems.
\end{itemize}

\section{Related Work}
\label{sec:related_work}

\subsection{Training Views and Gaussian Initialization}

Online 3DGS mapping optimizes the map from a growing stream of frames under a limited number of updates, so each system decides which frames to train on and how often. Most systems inherit this decision from the localization frontend: they optimize the map on a window of recent or covisible frames that the frontend keeps for localization and replay frames from earlier in the sequence~\cite{matsuki2024monogs,splatam2024,huang2024photoslam,splatslam2024,vigsslam}. Replaying past frames keeps mapped regions from being forgotten, as studied in continual learning for neural implicit maps~\cite{sucar2021imap,yan2021continual}. However, the frames kept for localization overlap little, leaving each region with too few views to constrain the many parameters of a 3DGS map, which therefore overfits to sparse inputs. We progressively admit the frames between them, which observe the same surfaces from additional viewpoints, into the training set.

Other methods allocate updates by scoring frames. Error-driven scheduling revisits past frames where the map fits poorly, measured by the last training loss of each frame~\cite{cartgs} or by rendering error accumulated over spatial regions~\cite{ellensohn2026eligsir}. These errors are recorded when frames are trained and do not reflect later updates of the map, and filtering out incoming frames that overlap the map~\cite{ellensohn2026eligsir} removes the overlapping views that a 3DGS map needs to constrain its parameters. Information-driven selection chooses views by the uncertainty of the map in each view, estimated from the scales and positional gradients of the Gaussians~\cite{lee2025novelview} or from the Fisher information of the map~\cite{jiang2024fisherrf}. Both evaluate every candidate view against the map, leaving fewer updates for optimization within the same real-time budget. Rather than spending the available time on scoring frames, we group frames by the supervision they provide and allocate updates to each group accordingly.

Each keyframe also inserts new Gaussians, and their number sets the model complexity that the few views of each region constrain. Most systems back-project keyframe depth and downsample it uniformly~\cite{matsuki2024monogs,splatslam2024,zhang2025hislam2}, or insert Gaussians where the rendered opacity is low or the rendering error is large~\cite{splatam2024,rtgslam2024}, placing as many Gaussians in smooth regions as in detailed ones. Image-guided sampling concentrates new Gaussians on image structure, measured by color gradients~\cite{yugay2023gaussianslam}, image contrast~\cite{wei2024gsfusion}, or edge responses~\cite{thomas2026gausslite,meuleman2025onthefly}, but still inserts a fixed number per frame or lets the number follow the absolute magnitude of the image response, which varies with the contrast of each scene. Neither adapts the number to how much detail a keyframe holds. We use image gradients to set both how many Gaussians each keyframe inserts relative to earlier keyframes and where they are placed within the image, so that the model complexity of the map follows the detail of the scene rather than a fixed budget or its contrast.

\begin{figure*}[!t]
  \centering
  \includegraphics[width=\textwidth]{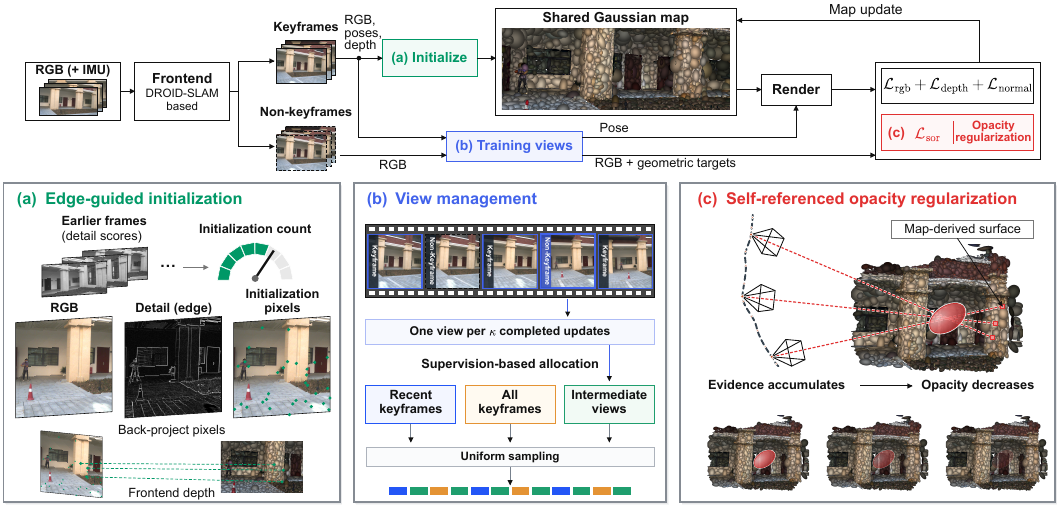}
  \caption{Overview of our method. (a) Edge-guided initialization controls the model complexity of the map by placing new Gaussians where image detail requires them. (b) View management admits non-keyframe images and allocates steps across recent keyframes, all keyframes, and non-keyframe images. (c) SOR suppresses the opacity of Gaussians that persistently contribute in front of the map-derived surface.}
  \label{fig:overview}
\end{figure*}

\subsection{Geometric Convergence and Floater Suppression}
Photometric reconstruction is geometrically under-constrained, since many opacity and depth configurations explain similar colors. Depth supervision constrains the rendering weights along each ray toward a given depth~\cite{deng2022dsnerf}, but in online mapping this depth comes from the frontend and is uncertain where viewpoints are few. Geometric regularization instead constrains the map itself: a distortion loss concentrates the rendering weights along each ray~\cite{mipnerf360}, and surface-oriented Gaussians are aligned with their normals and with planar depth that is consistent across neighboring views~\cite{huang20242dgs,chen2024pgsr}. These regularizers assume long offline optimization over all posed images, and their multi-view terms require additional renders. We complement frontend depth with a surface reference derived from the map itself, which accumulates over the views that training already renders.

Floaters can be handled by removing Gaussians or by lowering their opacity. Contribution-based pruning removes Gaussians whose blending weights across training views are small relative to their scale~\cite{fan2024trimgs}. Frequency-aware optimization emphasizes low-frequency structure to avoid floaters from poor initialization~\cite{wang2025efags}, and decoupled opacity separates the opacity used for geometry from that used for appearance~\cite{stablegs}. These offline methods judge each Gaussian over all posed images; online, a Gaussian is judged after only a few views, so removing it can discard surface that later views would support. Online mapping systems instead constrain opacity, forcing each Gaussian to be either opaque or nearly transparent~\cite{rtgslam2024} or penalizing the opacity of all Gaussians~\cite{cartgs}, but they constrain Gaussians on the surface as well as those in front of it. We lower only the opacity of Gaussians that persistently contribute in front of the map-derived surface, rather than removing them on the evidence of few views.

\section{Method}
\label{sec:method}

Given an image stream with the camera poses, keyframes, and keyframe depth estimated online by the frontend~\cite{vigsslam}, our goal is to optimize a 3DGS map that renders unseen viewpoints accurately and keeps its geometry consistent within a real-time budget.
Each optimization step renders one training image and updates the Gaussians visible in it with a base loss $\mathcal{L}_{\mathrm{base}}$, which combines L1 and D-SSIM photometric terms with depth and normal terms.
Because keyframes observe each region from few viewpoints, a map trained on them overfits to sparse inputs.
We address this from both sides (\cref{fig:overview}): edge-guided initialization controls the model complexity that each keyframe adds (\cref{ssec:pip_observations}), and view management adds non-keyframe images and divides the optimization steps among groups of images (\cref{ssec:pip_views}).
Finally, to complement uncertain frontend depth, Self-Referenced Opacity Regularization (SOR) suppresses floaters against a surface reference derived from the map itself (\cref{ssec:pip_optimization}).

\subsection{Edge-Guided Gaussian Initialization}
\label{ssec:pip_observations}
Each Gaussian adds parameters for the available views to constrain, so the number of Gaussians a keyframe adds sets the model complexity of the map. Gaussians placed in smooth regions add parameters with little detail to render, whereas too few Gaussians miss fine detail. We therefore decide how many Gaussians to initialize and where to place them from the image gradients of each incoming keyframe.

For each incoming keyframe \(k\), we compute the mean grayscale Sobel magnitude \(\bar{g}_k\) over pixels with valid depth estimates. We then calculate its empirical rank \(q_k \in (0,1)\) among the mean Sobel magnitudes of previously received keyframes. A higher rank indicates that the current image contains stronger gradients than earlier images.
We use this rank to adjust the number of initialized Gaussians:
\begin{equation}
    m_k=\min(1.5+2q_k,3.0),
\end{equation}
where \(m_k\) scales the base number of Gaussians obtained by downsampling the keyframe depth map. The resulting number of Gaussians, \(B_k\), is rounded to an integer and capped by the number of pixels with valid depth estimates. Using the rank rather than the absolute Sobel magnitude makes the number of initialized Gaussians less sensitive to differences in image-gradient magnitudes across scenes.

After determining \(B_k\), we select the pixels from which to initialize Gaussians. To place more Gaussians near edges and textured regions, we assign each pixel a sampling probability proportional to its Sobel magnitude:
\begin{equation}
    \begin{gathered}
    p_k(u)=\frac{g_k(u)+0.1\bar{g}_k}{\sum_{u'\in\Omega_k}(g_k(u')+0.1\bar{g}_k)},\\
    \bar{g}_k=\frac{1}{|\Omega_k|}\sum_{u\in\Omega_k}g_k(u),
    \end{gathered}
\end{equation}
where \(\Omega_k\) is the set of pixels with valid depth estimates, \(g_k(u)\) is the Sobel magnitude at pixel \(u\), and the $0.1\bar{g}_k$ allows pixels in smooth regions to be sampled as well. We then sample \(B_k\) pixels according to these probabilities without replacement.

Finally, we back-project the selected pixels into 3D using their estimated depths and camera poses to initialize Gaussian positions, and assign their RGB values as initial colors. When an existing map is available, we identify pixels already covered by the map using rendered opacity and depth consistency, and assign lower initial opacity to Gaussians initialized at these locations. We also limit Gaussian scales during optimization to prevent them from becoming excessively large.

\subsection{View Management: Admission and Allocation}  %
\label{ssec:pip_views}
Since each step updates only the Gaussians visible in its training image, the map is determined by which images are in the training set and how often each is drawn. As the stream grows while the number of steps stays limited, two decisions follow: which images to admit, and how to divide the steps among them.

\paragraph{Admitting non-keyframe images.} Keyframes observe each region from few viewpoints, which many configurations of the Gaussians reproduce equally well while rendering other viewpoints differently~\cite{zhu2024fsgs,sparsegs}. More updates on the same keyframes lower the training loss but do not resolve this ambiguity, whereas additional views of the same region constrain it. Such views are already in the stream: the non-keyframe images between consecutive keyframes. They refine the existing map without initializing new Gaussians; each takes its pose by interpolating its two neighboring keyframes and refining it with dense correspondences to them, and is supervised by its RGB and by depth warped from those keyframes where they agree.
Adding all of them at once would leave each image with too few updates. Instead, we admit one image for every \(\kappa\) completed optimization steps:
\begin{equation}
  |\mathcal{I}_t|\leq\left\lfloor\frac{S(t)}{\kappa}\right\rfloor,
  \label{eq:admission}
\end{equation}
where \(S(t)\) is the cumulative number of completed optimization steps up to time \(t\) and \(\mathcal{I}_t\) is the set of admitted non-keyframe images, so the training set grows only as fast as optimization proceeds. Since non-keyframe images receive a fixed share $n_{\mathcal{I}}/n$ of the steps (below), each is then updated about $\kappa n_{\mathcal{I}}/n$ times on average, however long the stream runs. We use \(\kappa=16\) and admit first the image temporally farthest from the training set, so that the admitted images cover the whole trajectory.

\paragraph{Allocating steps across groups.} The images provide different supervision as the map grows. Newly initialized Gaussians are seen mainly by the latest keyframes and need many updates soon; previously mapped regions need to be revisited, since updates for other images also move their Gaussians; and non-keyframe images constrain each region from additional viewpoints. We therefore draw from three groups: the recent keyframes $\mathcal{R}_t$, i.e., the latest $W$ keyframes; all keyframes $\mathcal{K}_t$, which revisit mapped regions with depth and normal supervision; and the admitted non-keyframe images $\mathcal{I}_t$. Each round of $n$ steps draws $n_{\mathcal{G}}$ images uniformly from each group $\mathcal{G}$, so that in expectation the steps descend
\begin{equation}
  \bar{\mathcal{L}}_t=\sum_{\mathcal{G}\in\{\mathcal{R}_t,\mathcal{K}_t,\mathcal{I}_t\}}\frac{n_{\mathcal{G}}}{n}\,\frac{1}{|\mathcal{G}|}\sum_{v\in\mathcal{G}}\mathcal{L}_v,
  \label{eq:allocation}
\end{equation}
where $\mathcal{L}_v$ is the mapping loss on image $v$. Each group keeps the weight $n_{\mathcal{G}}/n$ however large the map grows. A single pool over all training images $\mathcal{V}_t$ would instead weight each group by $|\mathcal{G}|/|\mathcal{V}_t|$; since $\mathcal{R}_t$ has a fixed size, the newest region would receive ever fewer steps as the stream grows. Within a group, we draw images uniformly; weighting images by update count or expected information gain~\cite{lee2025novelview} does not improve on this (\cref{ssec:view_analysis}). We use $n_{\mathcal{R}}{:}n_{\mathcal{K}}{:}n_{\mathcal{I}}=1{:}1{:}2$ with $n=12$, chosen empirically; details are provided in the supplementary material.

\subsection{Self-Referenced Opacity Regularization}
\label{ssec:pip_optimization}
\label{ssec:carve}

The depth term of $\mathcal{L}_{\mathrm{base}}$ relies on frontend depth, which is uncertain where viewpoints are few, so floaters can remain in front of the surface. As view management renders the map from many training views, the map accumulates its own evidence of where each surface lies. SOR uses this evidence as the reference: it measures how consistently each Gaussian contributes in front of the map-derived surface and suppresses the opacity of those that do so persistently.

\paragraph{Detecting Gaussians in Front of Surfaces.}
We use the weighted median depth as a surface reference, since averaging contributions at separated depths can place the mean between them. For ray $r$, let $z_{ri}$ be Gaussian $i$'s depth and $w_{ri}=T_{ri}\tilde{\alpha}_{ri}$ its blending weight, with transmittance $T_{ri}$ and effective opacity $\tilde{\alpha}_{ri}$. Following the renderer's front-to-back order, we obtain
\begin{equation}
 d_r^*=z_{r i^*},\qquad
 i^*=\min\left\{i:\sum_{j\le i}w_{rj}\ge A_r/2\right\},
 \label{eq:consensus_depth}
\end{equation}
where $A_r=\sum_iw_{ri}$. On the rays of training image $v$ with $A_r\ge A_{\min}$ and a finite reference, denoted by $\Omega_v$, we measure each Gaussian's front contribution:
\begin{equation}
 f_i=\sum_{r\in\Omega_v}w_{ri}\,\boldsymbol{1}[z_{ri}<(1-\rho)d_r^*],
 \label{eq:front_mass}
\end{equation}
where $\rho$ provides a relative depth margin.

To distinguish persistent inconsistency from occasional front contributions, we accumulate $F_i\leftarrow F_i+f_i$ and $V_i\leftarrow V_i+\sum_{r\in\Omega_v}w_{ri}$ across training renders. The front share $\phi_i=F_i/V_i$ determines
\begin{equation}
 \omega_i=\begin{cases}
 \operatorname{clip}\!\left(\dfrac{\phi_i-\tau_0}{\tau_1-\tau_0},0,1\right),&V_i>0,\\
 0,&V_i=0,
 \end{cases}
 \label{eq:persistence}
\end{equation}
where $\tau_0,\tau_1$ set the onset and saturation of suppression. Both the $F_i$ and $V_i$ accumulators are periodically decayed so that new observations can revise earlier evidence.

\paragraph{Opacity-Based Suppression.}
We penalize each Gaussian's current front contribution, weighted by $\omega_i$:
\begin{equation}
 \mathcal L_{\mathrm{sor}}=
 \frac{\lambda}{P}\sum_i \alpha_i
 \operatorname{sg}\!\left(\frac{\omega_i f_i}{\alpha_i}\right),
 \label{eq:consensus_carve}
\end{equation}
where $\lambda$ is the regularization weight, $\alpha_i$ is the opacity of Gaussian $i$, $P$ is the number of image pixels, and $\operatorname{sg}(\cdot)$ denotes the stop-gradient operator.
During each update, the coefficient inside $\operatorname{sg}(\cdot)$ is treated as constant, so each Gaussian's penalty contributes a gradient only to its own opacity.
This suppresses the inconsistent primitive itself rather than encouraging preceding Gaussians to obscure it through increased opacity.
Gradients through transmittance are retained for the base objective, which continues to refine appearance and geometry:
\begin{equation}
 \mathcal L_{\mathrm{map}}
 =\mathcal L_{\mathrm{base}}+\mathcal L_{\mathrm{sor}}.
 \label{eq:map_objective}
\end{equation}

Training reuses the current rendering buffers without additional renders or optimizer steps.
\definecolor{tablebest}{RGB}{190,35,35}
\definecolor{tablesecond}{RGB}{210,105,20}
\raggedbottom
\section{Experiments}
\label{sec:exp}
We first compare complete systems under real-time execution, then examine initialization, view management, and geometric refinement through controlled studies.

\subsection{Experimental Setup}
\label{ssec:setup}
\begin{figure*}[!t]
\centering
\captionof{table}{\textbf{Appearance and depth quality in native $1\times$ mapping.} Our method gives the best test-view appearance on every dataset and recovers the largest fraction of accurate depth. Depth is evaluated on TUM and FAST-LIVO2, which provide reference depth; because MAE ignores pixels where a map renders no depth, it should be read together with coverage. Values are means over five runs, with the \textcolor{tablebest}{\textbf{best}} and \textcolor{tablesecond}{\textbf{second-best}} per dataset highlighted.}
\label{tab:main}
\label{tab:rendering_comparison}
\label{tab:rendering_comparison_extended}
\label{tab:depth_comparison}
\begingroup
\footnotesize
\setlength{\tabcolsep}{2pt}
\renewcommand{\arraystretch}{0.93}
\begin{tabular*}{\textwidth}{@{\extracolsep{\fill}}llcrrrrrr@{}}
\toprule
Dataset & Method & \shortstack{Indoor/\\Outdoor} & PSNR $\uparrow$ & SSIM $\uparrow$ & LPIPS $\downarrow$ & \shortstack{Depth MAE\\(m) $\downarrow$} & \shortstack{Depth coverage\\(\%) $\uparrow$} & \shortstack{Within 5\,cm\\(\%) $\uparrow$} \\
\midrule
\multirow{6}{*}{Aria} & Photo-SLAM & \multirow{6}{*}{Indoor} & 17.10 & 0.686 & 0.653 & \textemdash & \textemdash & \textemdash \\
 & CaRtGS &  & 17.53 & \textcolor{tablesecond}{\textbf{0.689}} & 0.650 & \textemdash & \textemdash & \textemdash \\
 & VIGS-SLAM &  & 13.83 & 0.578 & 0.705 & \textemdash & \textemdash & \textemdash \\
 & HI-SLAM2 &  & \textcolor{tablesecond}{\textbf{18.09}} & 0.615 & \textcolor{tablesecond}{\textbf{0.626}} & \textemdash & \textemdash & \textemdash \\
 & Splat-SLAM &  & 12.93 & 0.298 & 0.796 & \textemdash & \textemdash & \textemdash \\
 & Ours &  & \textcolor{tablebest}{\textbf{20.92}} & \textcolor{tablebest}{\textbf{0.760}} & \textcolor{tablebest}{\textbf{0.463}} & \textemdash & \textemdash & \textemdash \\
\midrule
\multirow{6}{*}{TUM RGB-D} & Photo-SLAM & \multirow{6}{*}{Indoor} & 12.51 & \textcolor{tablesecond}{\textbf{0.532}} & 0.679 & 1.636 & \textcolor{tablebest}{\textbf{98.97}} & 14.94 \\
 & CaRtGS &  & 12.62 & 0.522 & 0.638 & 1.722 & 97.53 & 13.30 \\
 & VIGS-SLAM &  & 13.36 & 0.519 & 0.595 & \textcolor{tablebest}{\textbf{0.540}} & 92.37 & \textcolor{tablesecond}{\textbf{33.78}} \\
 & HI-SLAM2 &  & \textcolor{tablesecond}{\textbf{13.49}} & 0.481 & \textcolor{tablesecond}{\textbf{0.591}} & \textcolor{tablesecond}{\textbf{0.550}} & 91.00 & 31.86 \\
 & Splat-SLAM &  & 10.51 & 0.364 & 0.720 & 0.886 & 72.69 & 14.50 \\
 & Ours &  & \textcolor{tablebest}{\textbf{14.30}} & \textcolor{tablebest}{\textbf{0.551}} & \textcolor{tablebest}{\textbf{0.557}} & 0.572 & \textcolor{tablesecond}{\textbf{98.81}} & \textcolor{tablebest}{\textbf{37.53}} \\
\midrule
\multirow{6}{*}{FAST-LIVO2} & Photo-SLAM & \multirow{6}{*}{Outdoor} & 12.71 & 0.431 & 0.779 & 10.293 & 97.10 & 0.66 \\
 & CaRtGS &  & 13.34 & 0.437 & 0.735 & 11.830 & 97.96 & 0.61 \\
 & VIGS-SLAM &  & 16.50 & 0.476 & \textcolor{tablesecond}{\textbf{0.444}} & 0.862 & 90.47 & 19.86 \\
 & HI-SLAM2 &  & \textcolor{tablesecond}{\textbf{18.08}} & \textcolor{tablesecond}{\textbf{0.515}} & 0.456 & \textcolor{tablesecond}{\textbf{0.761}} & \textcolor{tablesecond}{\textbf{98.28}} & \textcolor{tablesecond}{\textbf{22.22}} \\
 & Splat-SLAM &  & 12.01 & 0.309 & 0.670 & 4.273 & 72.73 & 6.92 \\
 & Ours &  & \textcolor{tablebest}{\textbf{18.93}} & \textcolor{tablebest}{\textbf{0.523}} & \textcolor{tablebest}{\textbf{0.281}} & \textcolor{tablebest}{\textbf{0.558}} & \textcolor{tablebest}{\textbf{99.45}} & \textcolor{tablebest}{\textbf{23.04}} \\
\bottomrule
\end{tabular*}
\endgroup
\par\vspace{3pt}
  \centering
  \includegraphics[width=0.80\textwidth]{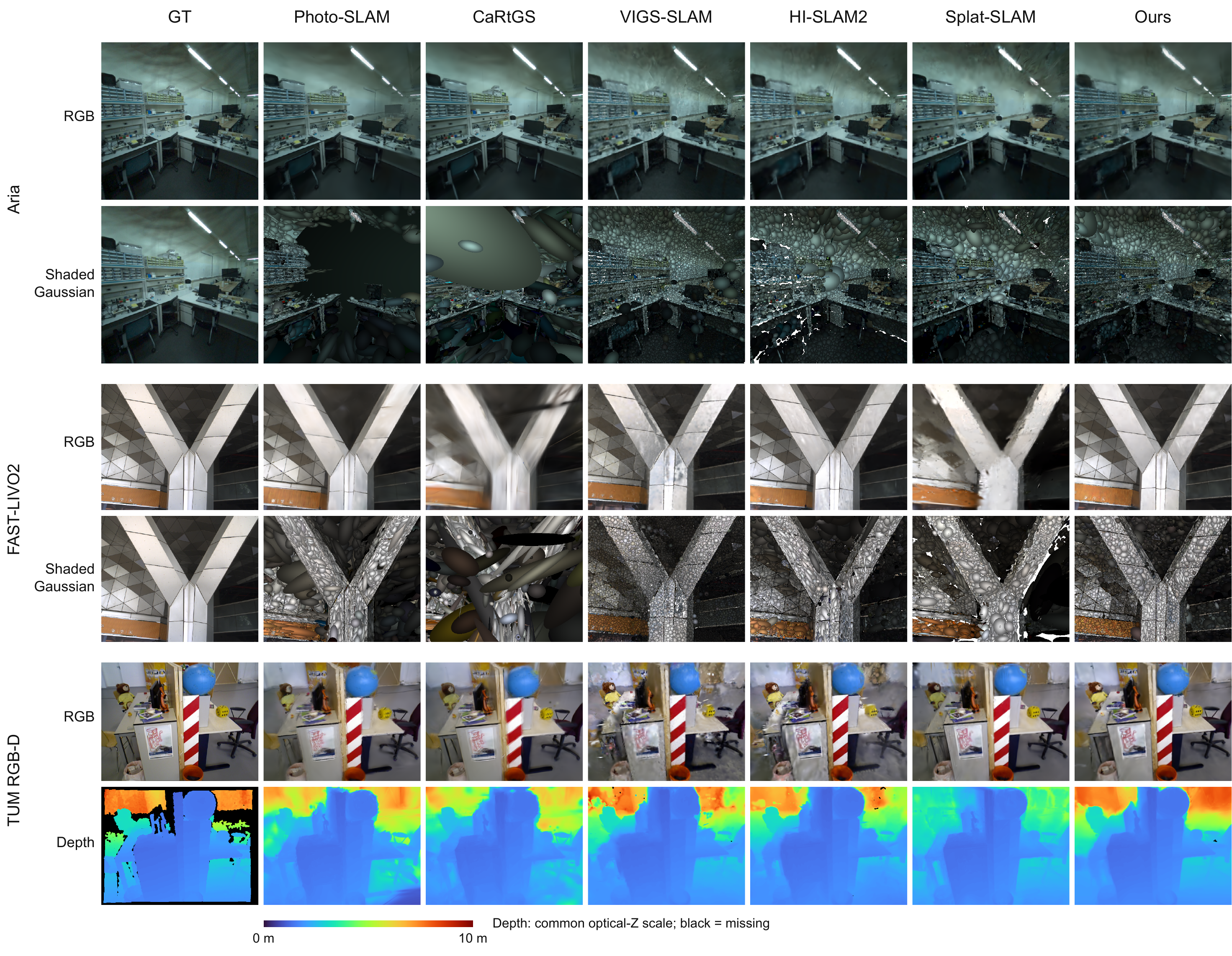}
  \captionof{figure}{\textbf{Similar images can hide different maps.} For Aria and FAST-LIVO2, rows show the rendered image and the Gaussians drawn as shaded ellipsoids; for TUM, which has dense reference depth, the second row shows rendered depth on a common metric scale. Photo-SLAM and CaRtGS render plausible images from oversized or missing Gaussians. On TUM, they and Splat-SLAM place the far wall too close, while VIGS-SLAM and HI-SLAM2 recover it only with visible image artifacts. Our method renders clean images from Gaussians that follow the scene structure.}
  \label{fig:rgb_comparison}
\end{figure*}
We compare Photo-SLAM~\cite{huang2024photoslam}, CaRtGS~\cite{cartgs}, VIGS-SLAM~\cite{vigsslam}, HI-SLAM2~\cite{zhang2025hislam2}, and Splat-SLAM~\cite{splatslam2024} on three datasets chosen to cover indoor and outdoor scenes. Aria sequences, which we recorded with head-worn Project Aria glasses~\cite{engel2023project}, add the fast rotations of natural head motion; TUM RGB-D~\cite{sturm2012tum} provides indoor scenes with registered depth, and FAST-LIVO2~\cite{zheng2025fastlivo2} provides outdoor scenes with LiDAR depth. UTMM~\cite{sun2024mm3dgs} adds RGB--IMU sequences with registered depth for the initialization and geometry studies (\cref{ssec:init_analysis,ssec:geometry_analysis}). All systems run on a desktop with an Intel Core i9-12900 CPU and a single NVIDIA GeForce RTX 5090 GPU.

Our work primarily focuses on mapping, so we evaluate the map that each system builds online rather than its trajectory. Inputs arrive at their original $1\times$ timestamps, and no optimizer update starts after the sequence ends; an update already in progress may finish. We hold out every fifth RGB image, together with the last one, as test views. These images are removed from the stream before it reaches the frontend, so no system tracks or trains on them. To render the test views, reference cameras are transformed into each map by one global Sim3 fitted to the training-camera poses.

Following 3DGS-based SLAM evaluations~\cite{matsuki2024monogs,splatam2024}, we report PSNR for pixel fidelity, SSIM for local structural similarity, and LPIPS for perceptual feature distance. Each of the three metrics is computed on every test view and averaged over the test views of a scene.

Image metrics measure appearance only, and depth error alone measures accuracy only where a map renders depth, so a map that leaves surfaces missing can still score well. We therefore compare rendered depth with reference depth using three metrics: depth MAE for accuracy, coverage for how completely the map covers the observed surfaces, and the fraction of pixels within 5\,cm, which requires both. Reference depth is registered sensor depth on TUM and projected LiDAR samples on FAST-LIVO2; Aria has none. Rendered depth is the opacity-normalized expected depth along the camera axis~\cite{huang20242dgs}, in metres, and a pixel is covered when the map renders a positive depth there with accumulated opacity of at least 0.5. Over the pixels with reference depth, MAE averages the absolute error between rendered and reference depth over covered pixels, coverage is the fraction of pixels that are covered, and within 5\,cm is the fraction of pixels that are covered and within 5\,cm of the reference depth. The floater fraction counts Gaussians whose centers lie closer than 75\% of the reference depth in at least 80\% of the test views observing them, among those observed in at least two; a center beyond 125\% of the reference depth is treated as occluded.

\subsection{Appearance and Geometry of Complete Systems}
\label{ssec:main}

In \cref{tab:main}, our method renders test views best on all three datasets in PSNR, SSIM, and LPIPS, and recovers the largest fraction of accurate depth (within 5\,cm) on both datasets with reference depth. On TUM, VIGS-SLAM and HI-SLAM2 reach a slightly lower MAE, but over fewer covered pixels.

Image quality alone, however, does not show whether a map has recovered the scene, so \cref{fig:rgb_comparison} also shows each map's Gaussians and, on TUM, its rendered depth. Baselines fail in two ways. Some render plausible images from oversized or missing Gaussians: they fit the training views without recovering the structure behind them, consistent with their lower test-view scores. Others recover the depth layout only at the cost of artifacts in the image. Our method avoids both, rendering clean images from Gaussians that follow the scene structure and matching the depth layout of the reference.
\subsection{Edge-Guided Initialization}
\label{ssec:abl}
\label{ssec:init_analysis}
\begin{table}[!t]
\centering
\caption{\textbf{Initialization ablation with controlled Gaussian counts.} Edge-guided initialization achieves the highest SSIM and lowest LPIPS on all eight UTMM sequences, and the highest mean PSNR overall. Results are averaged over three seeds.}
\label{tab:initialization}
\begingroup
\footnotesize
\setlength{\tabcolsep}{2pt}
\renewcommand{\arraystretch}{1.03}
\begin{tabular*}{\columnwidth}{@{\extracolsep{\fill}}lrrrr@{}}
\toprule
Initialization & \shortstack{Gaussians\\(k)} & PSNR $\uparrow$ & SSIM $\uparrow$ & LPIPS $\downarrow$ \\
\midrule
Uniform & 80.89 & 18.946 & \textcolor{tablesecond}{\textbf{0.6200}} & \textcolor{tablesecond}{\textbf{0.4601}} \\
On-the-Fly & 81.49 & \textcolor{tablesecond}{\textbf{19.143}} & 0.6149 & 0.5178 \\
Edge-guided (ours) & 80.66 & \textcolor{tablebest}{\textbf{19.252}} & \textcolor{tablebest}{\textbf{0.6335}} & \textcolor{tablebest}{\textbf{0.4246}} \\
\bottomrule
\end{tabular*}
\endgroup
\end{table}

To isolate initialization, all variants run in the same mapper and differ only in how new Gaussians are initialized. \Cref{tab:initialization} compares edge-guided initialization with uniform and On-the-Fly initialization~\cite{meuleman2025onthefly} at comparable Gaussian counts to examine whether allocating model complexity according to image detail improves reconstruction within a limited mapping budget. Our initialization method achieves the highest SSIM and lowest LPIPS on all eight UTMM sequences, together with the highest mean PSNR, using the available model complexity more effectively.
\subsection{View Management}
\label{ssec:view_analysis}
\begin{table}[t]
\centering
\includegraphics[width=\linewidth]{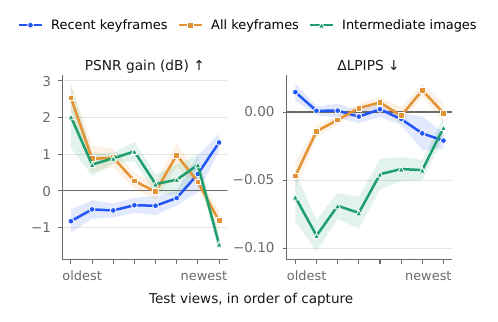}
\captionof{figure}{\textbf{Each group serves a different part of the stream.} Change in test-view quality when one group of \cref{ssec:pip_views} is removed and its updates go to the remaining groups, measured as our method minus the variant. Test views are split by capture time into eight groups of equal size; a positive PSNR gain or a negative $\Delta$LPIPS indicates that removing the group degrades quality. Recent keyframes improve the newest views most and all keyframes the oldest, while intermediate images lower LPIPS across the whole stream. Means over TUM RGB-D, Aria, and FAST-LIVO2; shading shows $\pm$1 standard error.}
\label{fig:view_groups}
\end{table}
\begin{table}[t]
\centering
\footnotesize
\setlength{\tabcolsep}{4pt}
\caption{\textbf{Scoring images within a group does not help in an online environment.} Differences from our method (variant $-$ ours) when images within each group of \cref{ssec:pip_views} are chosen by a score instead of uniformly; negative $\Delta$PSNR and positive $\Delta$LPIPS are worse. Means over TUM RGB-D, Aria, and FAST-LIVO2.}
\begin{tabular*}{\columnwidth}{@{\extracolsep{\fill}}lcccc@{}}
\toprule
\multirow{2}{*}{Sampling method} & \multicolumn{2}{c}{Whole map} & \multicolumn{2}{c}{$\Delta$PSNR per quarter} \\
\cmidrule(lr){2-3}\cmidrule(l){4-5}
 & $\Delta$PSNR & $\Delta$LPIPS & Oldest & Newest \\
\midrule
Update count & $+$0.06 & $-$0.002 & $-$0.51 & $+$0.39 \\
Information gain~\cite{lee2025novelview} & $-$0.37 & $+$0.031 & $+$0.19 & $-$0.11 \\
\bottomrule
\end{tabular*}
\label{tab:selection}
\end{table}
\begin{table}[!t]
\centering
\caption{\textbf{Geometry ablation on our system.} We replace our geometric treatment with map maintenance (MonoGS), depth regularization (DNGaussian), or detect-and-delete pruning (SparseGS), and compare our method without and with SOR. We use TUM RGB-D and UTMM, which provide reference depth for counting floaters. SOR leaves the fewest floaters on both datasets while keeping image quality. Values are means over sequences, with the \textcolor{tablebest}{\textbf{best}} and \textcolor{tablesecond}{\textbf{second-best}} per dataset highlighted.}
\label{tab:geometry_categories_seed0}
\begingroup
\footnotesize
\setlength{\tabcolsep}{2pt}
\renewcommand{\arraystretch}{1.0}
\def\B#1{\textcolor{tablebest}{\textbf{#1}}}\def\S#1{\textcolor{tablesecond}{\textbf{#1}}}
\begin{tabular*}{\columnwidth}{@{\extracolsep{\fill}}llrrrr@{}}
\toprule
Dataset & Method & PSNR $\uparrow$ & SSIM $\uparrow$ & LPIPS $\downarrow$ & \shortstack{Floaters\\(\%) $\downarrow$} \\
\midrule
\multirow{5}{*}{TUM RGB-D} & MonoGS & 14.15 & 0.483 & 0.572 & 13.67 \\
 & DNGaussian & 14.65 & 0.519 & 0.522 & 15.14 \\
 & SparseGS & \B{14.71} & 0.521 & \B{0.505} & 15.06 \\
 & w/o SOR & \S{14.70} & \S{0.525} & \S{0.506} & \S{13.43} \\
 & SOR & \B{14.71} & \B{0.528} & 0.509 & \B{12.52} \\
\midrule
\multirow{5}{*}{UTMM} & MonoGS & 14.46 & 0.437 & 0.549 & \S{1.09} \\
 & DNGaussian & \S{14.64} & \S{0.444} & 0.461 & 2.20 \\
 & SparseGS & 14.61 & 0.441 & \B{0.443} & 1.96 \\
 & w/o SOR & \B{14.72} & \B{0.445} & \S{0.456} & 1.23 \\
 & SOR & \B{14.72} & \B{0.445} & 0.459 & \B{0.92} \\
\bottomrule
\end{tabular*}
\endgroup
\end{table}
As \cref{fig:view_groups} shows, each group refines a part of the map that the others leave behind. Recent keyframes refine newly observed regions, which have had the least time to converge, and gain 1.3\,dB on the newest views. All keyframes keep refining regions that no longer appear among the recent keyframes, with a gain of 2.5\,dB on the oldest views. Intermediate images improve the appearance of the whole map rather than of one region.

These gains trade off: the updates of a removed group improve the views that the remaining groups serve. Without recent keyframes, whole-map PSNR even rises by 0.13\,dB while the newest views lose 1.3\,dB, so a whole-map average hides what each group contributes.

\Cref{tab:selection} replaces uniform sampling within each group with score-based choices, and neither score improves on uniform sampling. Weighting by update count moves quality from the oldest to the newest quarter with little change over the whole map, and selecting by expected information gain~\cite{lee2025novelview} degrades the whole map in an online environment.
\subsection{Self-Referenced Opacity Regularization}
\label{ssec:geometry_analysis}
\Cref{tab:geometry_categories_seed0} replaces our geometric treatment with representative alternatives while keeping the rest of our system unchanged under real-time execution: map maintenance (MonoGS~\cite{matsuki2024monogs}), depth regularization (DNGaussian~\cite{li2024dngaussian}), and detect-and-delete pruning (SparseGS~\cite{sparsegs}). SOR reduces the floater fraction from 13.4\% to 12.5\% on TUM and from 1.2\% to 0.9\% on UTMM, the lowest among all treatments, while keeping image quality on par with the best alternative.

\section{Conclusion}
\label{sec:conclusion}
We presented an online 3DGS mapping method built on a simple observation: SLAM selects keyframes for localization, but a 3DGS map, with its far higher model complexity, needs dense, overlapping views and overfits to sparse inputs when trained on keyframes alone.
The views it needs are already in the input stream; what limits the map is the real-time budget, in which every added view and Gaussian competes for the same updates.
Within this budget, view management progressively admits the non-keyframe images that SLAM leaves out and allocates updates across frame groups according to the supervision each provides, while edge-guided initialization lets the model complexity of the map follow the detail of the scene.
SOR complements uncertain frontend depth with a surface reference that the map accumulates from the views training already renders, suppressing floaters at no extra rendering cost.
On egocentric and public sequences at their original frame rate, our method improves the appearance of online maps without trading off their geometry.
A natural next step is mapping on the device itself, such as head-worn glasses, where tighter budgets make the choice of which images to train on and where to place Gaussians even more important.

\section*{LLM Usage Statement}
The authors designed the method and the experiments and wrote the draft of this paper.
We used LLM-based assistants (Claude, Anthropic; Codex and ChatGPT, OpenAI) to review the grammar and wording of the author-written text and suggest revisions, to assist in researching prior work, and to check citations and formatting.
They were also used to implement and run experiments, to manage and aggregate their results, and to produce figures under the authors' direction.
The authors reviewed and verified all content and take full responsibility for it.

{
    \small
    \bibliographystyle{ieeenat_fullname}
    \bibliography{rpm_packages/string-long,main}
}

\clearpage
\appendix
\setcounter{page}{1}
\renewcommand{\thepage}{S\arabic{page}}
\setcounter{table}{0}
\setcounter{figure}{0}
\setcounter{equation}{0}
\renewcommand{\thetable}{S\arabic{table}}
\renewcommand{\thefigure}{S\arabic{figure}}
\renewcommand{\theequation}{S\arabic{equation}}
\renewcommand{\theHtable}{supp.\arabic{table}}
\renewcommand{\theHfigure}{supp.\arabic{figure}}
\renewcommand{\theHequation}{supp.\arabic{equation}}
\providecommand{\authorfact}[1]{\textcolor{red}{[#1]}}
\definecolor{tablebest}{RGB}{190,35,35}
\definecolor{tablesecond}{RGB}{210,105,20}
\twocolumn[{
\begin{center}
\parbox{0.96\textwidth}{\centering\Large\bfseries View Management with Gaussian Placement and Opacity Regularization for Online 3DGS}\\[5pt]
{\large Supplementary Material}\\[9pt]
\end{center}
}]
\raggedbottom\makeatletter\@secpenalty=0\makeatother

This supplement contains the following sections.
\begin{list}{}{\setlength{\leftmargin}{1.1em}\setlength{\labelwidth}{0.8em}\setlength{\labelsep}{0.3em}\setlength{\itemsep}{1pt}\setlength{\parsep}{0pt}\setlength{\topsep}{2pt}}
\item[\textbf{\ref{sec:supp_impl}}] \textbf{Implementation Details}: settings of each component and resources.
\item[\textbf{\ref{sec:supp_eval}}] \textbf{Evaluation Details}: datasets, inputs, metrics, and failed runs.
\item[\textbf{\ref{sec:supp_init}}] \textbf{Initialization per Sequence}: main Table~2 on each UTMM sequence.
\item[\textbf{\ref{sec:supp_views}}] \textbf{Allocating Updates Among the Groups}: the comparison behind the 1:1:2 allocation of main Sec.~3.2.
\item[\textbf{\ref{sec:supp_sor}}] \textbf{SOR per Sequence}: main Table~4 on each sequence.
\item[\textbf{\ref{sec:supp_limits}}] \textbf{Limitations}.
\item[\textbf{\ref{sec:supp_tables}}] \textbf{Complete Systems per Sequence}: main Table~1 on each sequence and consistency across runs.
\end{list}
We also include a video that visually compares our maps with those of the baselines.

\section{Implementation Details}
\label{sec:supp_impl}
Our mapper runs concurrently with the frontend on the desktop of main Sec.~4.1 (Python 3.11, PyTorch 2.8, CUDA 12.8). Table~\ref{tab:supp_settings} lists its settings, which are the same for all sequences and datasets; only the use of IMU differs (Sec.~\ref{sec:supp_eval}).

\begin{table}[t]
\centering
\caption{\textbf{Settings of our mapper.} The same values are used for every sequence and dataset.}
\label{tab:supp_settings}
\begingroup
\footnotesize
\setlength{\tabcolsep}{3pt}
\renewcommand{\arraystretch}{1.02}
\begin{tabular*}{\columnwidth}{@{\extracolsep{\fill}}lr@{}}
\toprule
Setting & Value\\
\midrule
\multicolumn{2}{@{}l}{\emph{Optimization}}\\
Spherical-harmonic degree & 0\\
Position learning rate, initial / final & $1.6\!\times\!10^{-4}$ / $1.6\!\times\!10^{-6}$\\
Position learning-rate decay (updates) & 26,000\\
Color / opacity learning rate & 0.0025 / 0.05\\
Scale / rotation learning rate & 0.001 / 0.001\\
Adam $\beta_1,\beta_2,\epsilon$ & 0.9, 0.999, $10^{-15}$\\
Keyframe RGB / depth / normal weight & 0.95 / 0.25 / 0.05\\
Non-keyframe L1 / D-SSIM / depth weight & 0.8 / 0.2 / 0.25\\
\midrule
\multicolumn{2}{@{}l}{\emph{Initialization}}\\
First-keyframe multiplier $m_1$ & 2.5\\
Depth downsampling, first / later keyframes & 51.2 / 204.8\\
Initial opacity, covered / uncovered pixels & 0.02 / 0.5\\
Maximum scale (map units) & 0.5\\
\midrule
\multicolumn{2}{@{}l}{\emph{View management}}\\
Recent keyframes $W$ & 10\\
Updates per round $n_{\mathcal R}/n_{\mathcal K}/n_{\mathcal I}$ & 3/3/6\\
Admission interval $\kappa$ & 16\\
Images per update & 1\\
\midrule
\multicolumn{2}{@{}l}{\emph{SOR}}\\
Weight $\lambda$ & 3\\
Reference opacity $A_{\min}$ & 0.5\\
Front margin $\rho$ & 0.25\\
Onset / saturation $\tau_0$ / $\tau_1$ & 0.5 / 0.8\\
Decay of $F_i$ and $V_i$ every 300 updates & 0.5\\
\midrule
\multicolumn{2}{@{}l}{\emph{Opacity pruning}}\\
Interval (updates) & 300\\
Opacity threshold & 0.1\\
Protected latest insertion batches & 10\\
\bottomrule
\end{tabular*}
\endgroup
\end{table}
\begin{table}[t]
\centering
\caption{\textbf{Resources of our system.} Means over three runs, separate from those of Table~\ref{tab:supp_alloc}, with the 1:1:2 allocation on its sequences in native $1\times$ mapping. Selection is the time to choose the images of a round, per update.}
\label{tab:supp_resources}
\begingroup
\footnotesize
\setlength{\tabcolsep}{2pt}
\renewcommand{\arraystretch}{1.04}
\begin{tabular*}{\columnwidth}{@{\extracolsep{\fill}}lrrrrr@{}}
\toprule
Dataset & \shortstack{Stream\\(s)} & \shortstack{Completed\\updates} & \shortstack{Gaussians\\(k)} & \shortstack{Peak GPU\\memory (GiB)} & \shortstack{Selection\\(ms)} \\
\midrule
TUM RGB-D & 20.4 & 874 & 72.5 & 7.73 & 0.015 \\
Aria & 65.5 & 1655 & 132.8 & 8.79 & 0.032 \\
FAST-LIVO2 & 118.6 & 1754 & 108.6 & 9.38 & 0.028 \\
\bottomrule
\end{tabular*}
\endgroup
\end{table}

\paragraph{Supervision.} Keyframes are supervised by RGB, frontend depth, and depth-derived normals, and non-keyframe images by RGB and a depth target that projects the depth of the two neighboring keyframes into the image and averages them where they agree within 5\%. Table~\ref{tab:supp_settings} lists the weights of these terms in $\mathcal{L}_{\mathrm{base}}$.

\paragraph{Initialization.} The first keyframe uses $m_1=2.5$. For later keyframes, $q_k$ of main Sec.~3.1 is the mid-rank of the mean Sobel magnitude of keyframe $k$ among the previous keyframes,
\begin{equation}
 q_k=\frac{\ell_k+e_k/2+1/2}{N_{k-1}+1},
\end{equation}
where $N_{k-1}$ is the number of previous keyframes and $\ell_k$ and $e_k$ count those with a smaller and an equal mean. The base number of Gaussians is the number of pixels with valid depth in the keyframe depth map divided by 51.2 for the first keyframe and by 204.8 afterwards. New Gaussians start with opacity 0.5, or 0.02 where the map already covers the pixel, \ie, renders it with an accumulated opacity of at least 0.6 and the new point does not lie far in front of the rendered depth.

\paragraph{View management.} Each update renders one image and takes one optimizer step. Within a round, each group is sampled uniformly without replacement, and an image drawn as a recent keyframe is not drawn again from all keyframes. A non-keyframe image becomes eligible for admission once both neighboring keyframes have arrived; its pose is interpolated between them and refined by dense correspondence, with the IMU initializing its rotation when available.

\paragraph{SOR.} Since the coefficient of $\mathcal{L}_{\mathrm{sor}}$ is detached, the loss adds
\begin{equation}
 \frac{\partial\mathcal{L}_{\mathrm{sor}}}{\partial s_i}=\frac{\lambda}{P}\,\omega_i f_i\,(1-\alpha_i)
\end{equation}
to the gradient of the opacity logit $s_i$, with $\alpha_i=\sigma(s_i)$, and nothing to other parameters. The contributions $w_{ri}$ come from the training render, so SOR needs no extra render. $F_i$ and $V_i$ are halved every 300 updates.

\paragraph{Opacity pruning.} Every 300 updates, Gaussians with opacity below 0.1 are removed, except those of the ten latest insertion batches, which have received few updates so far.

\paragraph{Resources.} Table~\ref{tab:supp_resources} reports the memory and update counts of our system. Choosing the images of a round takes 0.015--0.032\,ms per update on average, below 0.1\% of the 23--68\,ms of stream time per update.

\section{Evaluation Details}
\label{sec:supp_eval}
\paragraph{Datasets.} Main Sec.~4.1 compares systems on three datasets that cover complementary conditions. The Aria recordings are egocentric: head-worn glasses add the fast rotations of natural head motion, but provide no reference depth. TUM RGB-D covers indoor scenes recorded with a handheld or robot-mounted RGB-D camera, including textureless and dynamic scenes, and its registered sensor depth serves as dense reference depth. FAST-LIVO2 covers large outdoor scenes, where projected LiDAR points serve as sparse reference depth. UTMM adds RGB--IMU sequences with registered depth and reference poses for the initialization and SOR studies (main Tables~2 and~4). TUM-01 to TUM-47 number the 47 TUM RGB-D sequences in alphabetical order of their names (TUM-01 is fr1/360 and TUM-47 is fr3/walking\_xyz). FAST-01 to FAST-05 are CBD\_Building\_01, CBD\_Building\_02, HKU\_Campus, Retail\_Street, and SYSU\_01, with reference cameras from the pseudo-ground-truth trajectories released by the dataset authors. UTMM sequences keep their dataset names.

\paragraph{Aria recordings.} We recorded eleven sequences of 65--243\,s with Project Aria glasses while walking through laboratories, lecture rooms, corridors, and stairwells of university buildings, with natural head rotation. The RGB camera runs at 20\,Hz (Aria-01 to Aria-06) or 10\,Hz (Aria-07 to Aria-11) and is rectified to $464\times464$ pixels. Reference cameras for the test views come from the closed-loop trajectories of Project Aria Machine Perception Services, which serve only for evaluation. Aria-05 and Aria-07, where some runs of CaRtGS failed, are excluded from main Table~1 (Table~\ref{tab:supp_availability}).

\begin{table}[t]
\centering
\caption{\textbf{Scored runs on all sequences.} Runs that produced a scored map, out of five per sequence. Main Table~1 averages each metric over the same sequences for every system: 9 Aria, 32 TUM RGB-D, and 5 FAST-LIVO2, and for depth 5 FAST-LIVO2 and 19 TUM RGB-D, since our evaluation did not associate the depth images of the other 13, all from the fr2 group, with their test views. A missing score excludes a sequence for all systems instead of counting as zero.}
\label{tab:supp_availability}
\begingroup
\footnotesize
\setlength{\tabcolsep}{2.3pt}
\renewcommand{\arraystretch}{1.04}
\begin{tabular*}{\columnwidth}{@{\extracolsep{\fill}}lrrr@{}}
\toprule
Method & Aria & TUM RGB-D & FAST-LIVO2 \\
\midrule
Photo-SLAM & 55/55 & 209/235 & 25/25 \\
CaRtGS & 52/55 & 201/235 & 25/25 \\
VIGS-SLAM & 55/55 & 235/235 & 25/25 \\
HI-SLAM2 & 55/55 & 230/235 & 25/25 \\
Splat-SLAM & 55/55 & 225/235 & 25/25 \\
Ours & 55/55 & 234/235 & 25/25 \\
\bottomrule
\end{tabular*}
\endgroup
\end{table}

\paragraph{Systems and inputs.} All systems receive monocular RGB at the original timestamps. Our method and VIGS-SLAM also use the IMU on Aria and FAST-LIVO2, and the other baselines run their monocular configurations. Sensor depth and LiDAR serve only for evaluation. Refinement that some baselines run after the stream, such as global bundle adjustment or map refinement, is disabled, since no optimizer update starts after the stream ends (main Sec.~4.1).

\begin{figure*}[t]
\centering
\footnotesize
\includegraphics{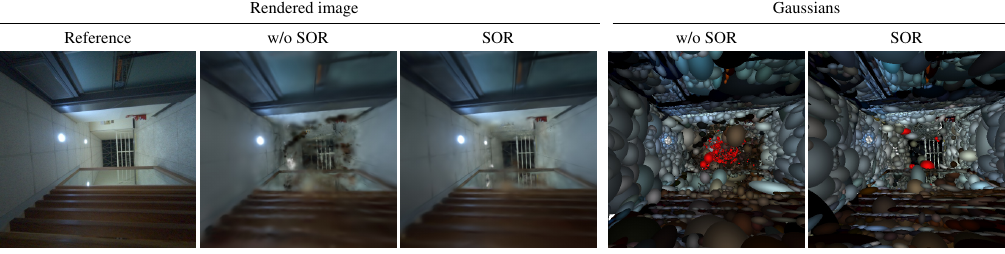}
\caption{\textbf{SOR on a test view of Aria-06.} Without SOR, floaters occlude the gate at the end of the corridor; with SOR, the gate is visible as in the reference. In the Gaussian panels, red marks the Gaussians that contribute mostly in front of the map-derived surface.}
\label{fig:supp_sor}
\end{figure*}

\paragraph{Metrics.} Depth metrics use reference depth between 0.1 and 50\,m; TUM RGB-D depth is used without hole filling, and FAST-LIVO2 uses only pixels with a projected LiDAR point. The floater counts in Table~\ref{tab:supp_sor} are the numerators of the floater fraction of main Sec.~4.1, whose denominator is the number of Gaussians observed in at least two test views.

\paragraph{Failed runs.} A run fails when the system crashes or ends without a map that can be evaluated; a run that loses tracking but saves a map is scored. Main Table~1 averages each dataset over the same sequences for all systems (Table~\ref{tab:supp_availability}). Most failures arise in the frontends. The ORB-SLAM3 frontend of Photo-SLAM and CaRtGS does not initialize in any of the five runs on the four textureless TUM RGB-D sequences (TUM-31, 32, 39, and 40) and loses tracking or produces invalid poses on several others, and on TUM-37, where the camera barely moves, the frontends produce too few keyframes for most systems to create Gaussians.

\setcounter{dbltopnumber}{3}
\begin{table*}[t]
\centering
\caption{\textbf{Edge-guided initialization gives the best SSIM and LPIPS on every sequence.} Main Table~2 per UTMM sequence, where U, O, and E denote uniform, On-the-Fly, and edge-guided (ours) initialization. The three run in the same mapper with the same updates, and their Gaussian counts match within 3.3\%. Values average three seeds, with the \textcolor{tablebest}{\textbf{best}} and \textcolor{tablesecond}{\textbf{second-best}} of each sequence highlighted.}
\label{tab:supp_init}
\begingroup
\footnotesize
\setlength{\tabcolsep}{2.4pt}
\renewcommand{\arraystretch}{1.15}
\begin{tabular*}{\textwidth}{@{\extracolsep{\fill}}lrrrrrrrrrrrr@{}}
\toprule
 & \multicolumn{3}{c}{Gaussians (k)} & \multicolumn{3}{c}{PSNR $\uparrow$} & \multicolumn{3}{c}{SSIM $\uparrow$} & \multicolumn{3}{c}{LPIPS $\downarrow$}\\
\cmidrule(lr){2-4}\cmidrule(lr){5-7}\cmidrule(lr){8-10}\cmidrule(lr){11-13}
Sequence & U & O & E & U & O & E & U & O & E & U & O & E\\
\midrule
square-1 & 119.74 & 119.80 & 119.14 & \textcolor{tablesecond}{\textbf{18.544}} & 18.410 & \textcolor{tablebest}{\textbf{18.752}} & \textcolor{tablesecond}{\textbf{0.5790}} & 0.5729 & \textcolor{tablebest}{\textbf{0.5901}} & \textcolor{tablesecond}{\textbf{0.4964}} & 0.5326 & \textcolor{tablebest}{\textbf{0.4600}}\\
square-2 & 117.09 & 118.05 & 117.04 & 18.908 & \textcolor{tablesecond}{\textbf{18.929}} & \textcolor{tablebest}{\textbf{19.109}} & \textcolor{tablesecond}{\textbf{0.6040}} & 0.5966 & \textcolor{tablebest}{\textbf{0.6161}} & \textcolor{tablesecond}{\textbf{0.4833}} & 0.5376 & \textcolor{tablebest}{\textbf{0.4475}}\\
ego-centric-1 & 100.60 & 102.34 & 100.09 & \textcolor{tablesecond}{\textbf{19.315}} & 19.181 & \textcolor{tablebest}{\textbf{19.520}} & \textcolor{tablesecond}{\textbf{0.6393}} & 0.6359 & \textcolor{tablebest}{\textbf{0.6557}} & \textcolor{tablesecond}{\textbf{0.4386}} & 0.5023 & \textcolor{tablebest}{\textbf{0.4036}}\\
ego-centric-2 & 107.73 & 108.00 & 107.30 & \textcolor{tablesecond}{\textbf{19.714}} & 19.664 & \textcolor{tablebest}{\textbf{19.968}} & \textcolor{tablesecond}{\textbf{0.6396}} & 0.6381 & \textcolor{tablebest}{\textbf{0.6521}} & \textcolor{tablesecond}{\textbf{0.4309}} & 0.4867 & \textcolor{tablebest}{\textbf{0.4027}}\\
ego-drive & 119.48 & 120.69 & 119.28 & 18.220 & \textcolor{tablebest}{\textbf{18.341}} & \textcolor{tablesecond}{\textbf{18.301}} & 0.5521 & \textcolor{tablesecond}{\textbf{0.5558}} & \textcolor{tablebest}{\textbf{0.5581}} & \textcolor{tablesecond}{\textbf{0.5338}} & 0.5584 & \textcolor{tablebest}{\textbf{0.4991}}\\
fast-straight & 33.22 & 33.22 & 33.23 & 19.454 & \textcolor{tablesecond}{\textbf{19.751}} & \textcolor{tablebest}{\textbf{20.149}} & \textcolor{tablesecond}{\textbf{0.6434}} & 0.6208 & \textcolor{tablebest}{\textbf{0.6641}} & \textcolor{tablesecond}{\textbf{0.4738}} & 0.5860 & \textcolor{tablebest}{\textbf{0.4317}}\\
slow-straight-1 & 19.66 & 20.06 & 19.67 & 17.338 & \textcolor{tablebest}{\textbf{18.093}} & \textcolor{tablesecond}{\textbf{17.689}} & 0.6174 & \textcolor{tablesecond}{\textbf{0.6273}} & \textcolor{tablebest}{\textbf{0.6339}} & \textcolor{tablesecond}{\textbf{0.4359}} & 0.4942 & \textcolor{tablebest}{\textbf{0.3993}}\\
slow-straight-2 & 29.54 & 29.75 & 29.56 & 20.076 & \textcolor{tablebest}{\textbf{20.778}} & \textcolor{tablesecond}{\textbf{20.531}} & \textcolor{tablesecond}{\textbf{0.6852}} & 0.6717 & \textcolor{tablebest}{\textbf{0.6978}} & \textcolor{tablesecond}{\textbf{0.3885}} & 0.4444 & \textcolor{tablebest}{\textbf{0.3531}}\\
\midrule
Mean & 80.89 & 81.49 & 80.66 & 18.946 & \textcolor{tablesecond}{\textbf{19.143}} & \textcolor{tablebest}{\textbf{19.252}} & \textcolor{tablesecond}{\textbf{0.6200}} & 0.6149 & \textcolor{tablebest}{\textbf{0.6335}} & \textcolor{tablesecond}{\textbf{0.4601}} & 0.5178 & \textcolor{tablebest}{\textbf{0.4246}}\\
\bottomrule
\end{tabular*}
\endgroup
\end{table*}
\begin{table*}[t]
\centering
\caption{\textbf{1:1:2 gives the lowest LPIPS of the five allocations.} Each row divides the 12 updates of a round among recent keyframes, all keyframes, and non-keyframe images ($n_{\mathcal R}/n_{\mathcal K}/n_{\mathcal I}$); ours is 3/3/6, i.e., 1:1:2. Differences are from ours, averaged over three runs on one sequence per dataset; a positive $\Delta$PSNR or a negative $\Delta$LPIPS would favor the alternative. Parentheses give each difference relative to the value of ours in the same column. The last three columns count the updates completed within each stream.}
\label{tab:supp_alloc}
\begingroup
\footnotesize
\setlength{\tabcolsep}{4pt}
\renewcommand{\arraystretch}{1.15}
\begin{tabular*}{\textwidth}{@{\extracolsep{\fill}}lrrrrrrrr@{}}
\toprule
 & \multicolumn{4}{c}{$\Delta$PSNR (dB)} & & \multicolumn{3}{c}{Completed updates}\\
\cmidrule(lr){2-5}\cmidrule(lr){7-9}
$n_{\mathcal R}/n_{\mathcal K}/n_{\mathcal I}$ & TUM & Aria & FAST & Mean & Mean $\Delta$LPIPS & TUM & Aria & FAST\\
\midrule
3/3/6 (ours) & -- & -- & -- & -- & -- & 877 & 1697 & 1720\\
1/7/4 & $-$0.003 ($-$0.02\%) & $+$0.054 ($+$0.21\%) & $+$0.080 ($+$0.35\%) & $+$0.043 ($+$0.19\%) & $+$0.0222 ($+$7.92\%) & 902 & 1680 & 1746\\
5/4/3 & $-$0.001 ($-$0.01\%) & $-$0.235 ($-$0.92\%) & $-$0.694 ($-$3.03\%) & $-$0.310 ($-$1.36\%) & $+$0.0307 ($+$10.92\%) & 906 & 1710 & 1909\\
6/3/3 & $-$0.258 ($-$1.29\%) & $-$0.624 ($-$2.46\%) & $-$0.997 ($-$4.36\%) & $-$0.626 ($-$2.75\%) & $+$0.0324 ($+$11.56\%) & 905 & 1961 & 1839\\
8/4/0 & $-$0.624 ($-$3.12\%) & $-$0.936 ($-$3.69\%) & $-$1.372 ($-$6.00\%) & $-$0.977 ($-$4.30\%) & $+$0.0549 ($+$19.54\%) & 1139 & 3157 & 2843\\
\bottomrule
\end{tabular*}
\endgroup
\end{table*}

\section{Initialization per Sequence}
\label{sec:supp_init}
Table~\ref{tab:supp_init} lists main Table~2 per UTMM sequence. So that the three variants differ only in initialization, they receive identical inputs: the same keyframes and keyframe depth from one frontend run, the reference poses of UTMM, and a fixed budget of two updates per image of the stream, spent on keyframes. The number of Gaussians that uniform and On-the-Fly initialization insert is calibrated per sequence on the first of the three seeds, so that their final counts match ours within 3.3\% for every seed. Edge-guided initialization gives the best SSIM and LPIPS on all eight sequences and the best PSNR on five; on the other three, On-the-Fly initialization has a higher PSNR but an LPIPS higher by 0.059--0.095.

\section{Allocating Updates Among the Groups}
\label{sec:supp_views}
Main Sec.~3.2 divides the $n=12$ updates of each round among recent keyframes, all keyframes, and non-keyframe images as 1:1:2, \ie, 3/3/6 images. Table~\ref{tab:supp_alloc} compares it in native $1\times$ mapping, on one sequence per dataset with three runs each, with allocations that shift updates to all keyframes (1/7/4), to recent keyframes (5/4/3 and 6/3/3), and away from non-keyframe images entirely (8/4/0). LPIPS is higher than with 1:1:2 in all 36 comparisons (four alternatives, three sequences, three runs). Mean PSNR falls by 0.31, 0.63, and 0.98\,dB as updates move from non-keyframe images to recent keyframes (5/4/3, 6/3/3, and 8/4/0), and 1/7/4 stays on par in PSNR ($+$0.04\,dB) while raising LPIPS by 0.022. Without non-keyframe images, the mapper completes 1.3--1.9$\times$ as many updates within the same stream yet gives the lowest quality: updates on non-keyframe images improve the map more than additional updates on keyframes.

\begin{table}[!t]
\centering
\caption{\textbf{SOR reduces floaters on most sequences.} Floaters (main Sec.~4.1) per sequence of main Table~4 without and with SOR. $\Delta$PSNR and $\Delta$LPIPS compare SOR with w/o SOR. Counts are means over runs; the last rows give the floater fractions of main Table~4.}
\label{tab:supp_sor}
\begingroup
\footnotesize
\setlength{\tabcolsep}{1.5pt}
\renewcommand{\arraystretch}{1.02}
\begin{tabular*}{\columnwidth}{@{\extracolsep{\fill}}lrrrr@{}}
\toprule
Sequence & w/o SOR & SOR & $\Delta$PSNR & $\Delta$LPIPS \\
\midrule
TUM-01 & 103145 & 82725 & $+$0.837 & $-$0.0612 \\
TUM-02 & 571 & 599 & $+$0.008 & $+$0.0039 \\
TUM-03 & 3056 & 5700 & $-$0.217 & $+$0.0031 \\
TUM-04 & 1053 & 2258 & $-$0.458 & $+$0.0204 \\
TUM-05 & 2201 & 2119 & $-$0.026 & $+$0.0079 \\
TUM-06 & 51069 & 45218 & $-$0.283 & $+$0.0125 \\
TUM-07 & 15738 & 15005 & $+$0.050 & $+$0.0040 \\
TUM-08 & 1074 & 969 & $-$0.041 & $+$0.0038 \\
TUM-13 & 28 & 15 & $-$0.042 & $+$0.0079 \\
TUM-14 & 129 & 111 & $-$0.011 & $+$0.0116 \\
TUM-15 & 454 & 251 & $+$0.091 & $-$0.0016 \\
TUM-16 & 738 & 637 & $-$0.272 & $+$0.0186 \\
TUM-18 & 15790 & 17110 & $+$0.454 & $-$0.0109 \\
TUM-20 & 985 & 666 & $-$0.059 & $+$0.0115 \\
TUM-22 & 263 & 163 & $-$0.012 & $-$0.0087 \\
TUM-26 & 1908 & 963 & $+$0.128 & $+$0.0233 \\
TUM-27 & 317 & 296 & $+$0.000 & $-$0.0074 \\
TUM-29 & 1179 & 1263 & $-$0.054 & $+$0.0137 \\
TUM-30 & 562 & 459 & $-$0.041 & $+$0.0093 \\
TUM-34 & 0 & 0 & $-$0.062 & $+$0.0067 \\
TUM-35 & 1199 & 1063 & $+$0.037 & $-$0.0025 \\
TUM-36 & 11616 & 9490 & $+$0.031 & $+$0.0020 \\
TUM-38 & 326 & 319 & $-$0.013 & $+$0.0028 \\
TUM-41 & 50 & 56 & $-$0.087 & $+$0.0188 \\
TUM-42 & 353 & 330 & $-$0.060 & $+$0.0065 \\
TUM-45 & 3423 & 2852 & $+$0.270 & $-$0.0083 \\
TUM-47 & 261 & 193 & $+$0.098 & $+$0.0120 \\
ego-centric-1 & 606 & 541 & $+$0.007 & $+$0.0007 \\
ego-centric-2 & 354 & 248 & $-$0.017 & $+$0.0023 \\
ego-drive & 1204 & 781 & $+$0.041 & $+$0.0089 \\
square-1 & 1240 & 939 & $-$0.014 & $+$0.0012 \\
\midrule
TUM RGB-D (\%) & 13.43 & 12.52 & $+$0.01 & $+$0.003 \\
UTMM (\%) & 1.23 & 0.92 & $\pm$0.00 & $+$0.003 \\
\bottomrule
\end{tabular*}
\endgroup
\end{table}

\section{SOR per Sequence}
\label{sec:supp_sor}
Table~\ref{tab:supp_sor} lists main Table~4 per sequence for our method without and with SOR. This study uses 31 TUM RGB-D sequences fixed before it ran and keeps the 27 on which all five variants of main Table~4 completed the same runs. SOR reduces floaters on 20 of 27 TUM RGB-D sequences and on all four UTMM sequences, and it keeps image quality: the mean PSNR and LPIPS of each dataset change by at most 0.01\,dB and 0.003. Fig.~\ref{fig:supp_sor} shows an example on Aria-06.

\section{Limitations}
\label{sec:supp_limits}
SOR complements uncertain frontend depth with a surface reference that the map accumulates from its training views. This reference, however, comes from a map that is itself fitted to frontend depth: where frontend depth departs far from the true depth over many views, the map places the surface itself at the wrong depth, and SOR, which suppresses only Gaussians in front of this surface, does not correct it. Our method thus still relies on the depth accuracy of its frontend.

\setcounter{topnumber}{4}\setcounter{bottomnumber}{2}\setcounter{totalnumber}{6}\renewcommand{\topfraction}{0.95}\renewcommand{\bottomfraction}{0.9}\renewcommand{\textfraction}{0.05}\renewcommand{\floatpagefraction}{0.85}\setlength{\intextsep}{8pt plus 2pt minus 2pt}\setlength{\floatsep}{10pt plus 2pt minus 2pt}
\section{Complete Systems per Sequence}
\label{sec:supp_tables}
Tables~\ref{tab:supp_system_aria_psnr}--\ref{tab:supp_system_fast_livo2_within5cm} list main Table~1 per sequence as means over the five runs, with the \textcolor{tablebest}{\textbf{best}} and \textcolor{tablesecond}{\textbf{second-best}} value of each row highlighted. In each of the five runs, our method gives the best dataset mean in PSNR, SSIM, and LPIPS on all three datasets and in depth within 5\,cm on TUM RGB-D and FAST-LIVO2; the only exception is SSIM on FAST-LIVO2 in one run, where it ranks second.

\FloatBarrier
\subsection{Aria}
\label{sec:supp_tables_aria}
\begin{table}[!htb]
\centering
\caption{\textbf{Aria: PSNR per sequence.}}
\label{tab:supp_system_aria_psnr}
\begingroup
\footnotesize
\setlength{\tabcolsep}{2.3pt}
\renewcommand{\arraystretch}{1.04}
\begin{tabular*}{\columnwidth}{@{\extracolsep{\fill}}lrrrrrr@{}}
\toprule
Sequence & \shortstack{Photo-\\SLAM} & CaRtGS & \shortstack{VIGS-\\SLAM} & \shortstack{HI-\\SLAM2} & \shortstack{Splat-\\SLAM} & Ours \\
\midrule
Aria-01 & 21.01 & \textcolor{tablesecond}{\textbf{21.28}} & 17.87 & 20.32 & 14.75 & \textcolor{tablebest}{\textbf{23.16}} \\
Aria-02 & 21.13 & \textcolor{tablesecond}{\textbf{22.59}} & 17.77 & 19.31 & 13.15 & \textcolor{tablebest}{\textbf{23.58}} \\
Aria-03 & 18.45 & \textcolor{tablebest}{\textbf{19.70}} & 10.70 & 16.93 & 12.15 & \textcolor{tablesecond}{\textbf{18.71}} \\
Aria-04 & 15.94 & 16.66 & 13.95 & \textcolor{tablesecond}{\textbf{16.75}} & 12.34 & \textcolor{tablebest}{\textbf{19.43}} \\
Aria-06 & 14.97 & 16.10 & 16.85 & \textcolor{tablesecond}{\textbf{18.66}} & 11.96 & \textcolor{tablebest}{\textbf{21.12}} \\
Aria-08 & 16.67 & 16.49 & 7.41 & \textcolor{tablesecond}{\textbf{16.86}} & 11.40 & \textcolor{tablebest}{\textbf{18.51}} \\
Aria-09 & 15.94 & 15.78 & 13.04 & \textcolor{tablesecond}{\textbf{19.88}} & 13.84 & \textcolor{tablebest}{\textbf{22.36}} \\
Aria-10 & 15.47 & 14.97 & 12.65 & \textcolor{tablesecond}{\textbf{16.61}} & 13.34 & \textcolor{tablebest}{\textbf{19.66}} \\
Aria-11 & 14.35 & 14.22 & 14.25 & \textcolor{tablesecond}{\textbf{17.48}} & 13.47 & \textcolor{tablebest}{\textbf{21.77}} \\
\bottomrule
\end{tabular*}
\endgroup
\end{table}
\begin{table}[!htb]
\centering
\caption{\textbf{Aria: SSIM per sequence.}}
\label{tab:supp_system_aria_ssim}
\begingroup
\footnotesize
\setlength{\tabcolsep}{2.3pt}
\renewcommand{\arraystretch}{1.04}
\begin{tabular*}{\columnwidth}{@{\extracolsep{\fill}}lrrrrrr@{}}
\toprule
Sequence & \shortstack{Photo-\\SLAM} & CaRtGS & \shortstack{VIGS-\\SLAM} & \shortstack{HI-\\SLAM2} & \shortstack{Splat-\\SLAM} & Ours \\
\midrule
Aria-01 & 0.691 & \textcolor{tablesecond}{\textbf{0.703}} & 0.602 & 0.682 & 0.439 & \textcolor{tablebest}{\textbf{0.756}} \\
Aria-02 & 0.715 & \textcolor{tablesecond}{\textbf{0.737}} & 0.636 & 0.660 & 0.353 & \textcolor{tablebest}{\textbf{0.800}} \\
Aria-03 & \textcolor{tablesecond}{\textbf{0.794}} & \textcolor{tablebest}{\textbf{0.803}} & 0.578 & 0.500 & 0.204 & 0.765 \\
Aria-04 & 0.684 & \textcolor{tablesecond}{\textbf{0.697}} & 0.562 & 0.647 & 0.262 & \textcolor{tablebest}{\textbf{0.705}} \\
Aria-06 & 0.623 & \textcolor{tablesecond}{\textbf{0.679}} & 0.639 & 0.669 & 0.203 & \textcolor{tablebest}{\textbf{0.756}} \\
Aria-08 & \textcolor{tablesecond}{\textbf{0.721}} & 0.685 & 0.514 & 0.566 & 0.098 & \textcolor{tablebest}{\textbf{0.780}} \\
Aria-09 & 0.680 & 0.680 & 0.551 & \textcolor{tablesecond}{\textbf{0.717}} & 0.399 & \textcolor{tablebest}{\textbf{0.781}} \\
Aria-10 & \textcolor{tablesecond}{\textbf{0.661}} & 0.641 & 0.561 & 0.486 & 0.319 & \textcolor{tablebest}{\textbf{0.727}} \\
Aria-11 & 0.609 & 0.580 & 0.560 & \textcolor{tablesecond}{\textbf{0.613}} & 0.404 & \textcolor{tablebest}{\textbf{0.770}} \\
\bottomrule
\end{tabular*}
\endgroup
\end{table}
\begin{table}[!htb]
\centering
\caption{\textbf{Aria: LPIPS per sequence.}}
\label{tab:supp_system_aria_lpips}
\begingroup
\footnotesize
\setlength{\tabcolsep}{2.3pt}
\renewcommand{\arraystretch}{1.04}
\begin{tabular*}{\columnwidth}{@{\extracolsep{\fill}}lrrrrrr@{}}
\toprule
Sequence & \shortstack{Photo-\\SLAM} & CaRtGS & \shortstack{VIGS-\\SLAM} & \shortstack{HI-\\SLAM2} & \shortstack{Splat-\\SLAM} & Ours \\
\midrule
Aria-01 & \textcolor{tablesecond}{\textbf{0.349}} & 0.383 & 0.562 & 0.470 & 0.702 & \textcolor{tablebest}{\textbf{0.327}} \\
Aria-02 & 0.376 & \textcolor{tablebest}{\textbf{0.318}} & 0.551 & 0.502 & 0.743 & \textcolor{tablesecond}{\textbf{0.331}} \\
Aria-03 & 0.641 & \textcolor{tablesecond}{\textbf{0.603}} & 0.776 & 0.685 & 0.812 & \textcolor{tablebest}{\textbf{0.576}} \\
Aria-04 & 0.735 & 0.745 & 0.730 & \textcolor{tablesecond}{\textbf{0.684}} & 0.818 & \textcolor{tablebest}{\textbf{0.551}} \\
Aria-06 & 0.786 & 0.770 & 0.699 & \textcolor{tablesecond}{\textbf{0.688}} & 0.850 & \textcolor{tablebest}{\textbf{0.510}} \\
Aria-08 & 0.690 & 0.705 & 0.873 & \textcolor{tablesecond}{\textbf{0.687}} & 0.880 & \textcolor{tablebest}{\textbf{0.580}} \\
Aria-09 & 0.720 & 0.744 & 0.721 & \textcolor{tablesecond}{\textbf{0.586}} & 0.784 & \textcolor{tablebest}{\textbf{0.402}} \\
Aria-10 & 0.760 & 0.769 & 0.727 & \textcolor{tablesecond}{\textbf{0.675}} & 0.799 & \textcolor{tablebest}{\textbf{0.490}} \\
Aria-11 & 0.821 & 0.814 & 0.709 & \textcolor{tablesecond}{\textbf{0.657}} & 0.774 & \textcolor{tablebest}{\textbf{0.402}} \\
\bottomrule
\end{tabular*}
\endgroup
\end{table}

\FloatBarrier
\subsection{TUM RGB-D}
\label{sec:supp_tables_tum_rgbd}
\begin{table}[!htb]
\centering
\caption{\textbf{TUM RGB-D: PSNR per sequence.}}
\label{tab:supp_system_tum_rgbd_psnr}
\begingroup
\footnotesize
\setlength{\tabcolsep}{2.3pt}
\renewcommand{\arraystretch}{1.04}
\begin{tabular*}{\columnwidth}{@{\extracolsep{\fill}}lrrrrrr@{}}
\toprule
Sequence & \shortstack{Photo-\\SLAM} & CaRtGS & \shortstack{VIGS-\\SLAM} & \shortstack{HI-\\SLAM2} & \shortstack{Splat-\\SLAM} & Ours \\
\midrule
TUM-03 & 10.16 & 11.08 & 10.86 & \textcolor{tablesecond}{\textbf{11.58}} & 5.10 & \textcolor{tablebest}{\textbf{13.81}} \\
TUM-05 & 10.79 & 10.09 & \textcolor{tablesecond}{\textbf{11.02}} & 8.39 & 7.95 & \textcolor{tablebest}{\textbf{12.85}} \\
TUM-07 & \textcolor{tablebest}{\textbf{8.94}} & 7.73 & 4.22 & 4.52 & 4.04 & \textcolor{tablesecond}{\textbf{8.60}} \\
TUM-08 & 8.87 & 9.34 & \textcolor{tablesecond}{\textbf{11.34}} & 11.04 & 8.65 & \textcolor{tablebest}{\textbf{12.61}} \\
TUM-09 & 11.66 & 12.00 & 13.22 & \textcolor{tablesecond}{\textbf{14.20}} & 7.26 & \textcolor{tablebest}{\textbf{15.11}} \\
TUM-10 & \textcolor{tablesecond}{\textbf{12.09}} & 10.49 & 6.59 & \textcolor{tablebest}{\textbf{13.30}} & 10.74 & 4.94 \\
TUM-11 & \textcolor{tablesecond}{\textbf{9.56}} & 8.31 & 4.83 & 8.69 & \textcolor{tablebest}{\textbf{10.31}} & 5.93 \\
TUM-13 & 13.68 & 14.23 & \textcolor{tablesecond}{\textbf{15.70}} & 14.94 & 12.19 & \textcolor{tablebest}{\textbf{16.06}} \\
TUM-14 & 9.87 & 12.57 & 14.14 & \textcolor{tablesecond}{\textbf{14.18}} & 10.26 & \textcolor{tablebest}{\textbf{14.68}} \\
TUM-15 & 9.41 & 9.40 & \textcolor{tablesecond}{\textbf{15.28}} & 14.81 & 11.26 & \textcolor{tablebest}{\textbf{15.46}} \\
TUM-16 & 13.08 & 13.75 & 15.48 & \textcolor{tablesecond}{\textbf{15.93}} & 12.50 & \textcolor{tablebest}{\textbf{16.29}} \\
TUM-17 & 10.93 & 10.18 & 14.86 & \textcolor{tablesecond}{\textbf{15.37}} & 12.53 & \textcolor{tablebest}{\textbf{15.87}} \\
TUM-18 & \textcolor{tablesecond}{\textbf{11.07}} & 10.22 & 6.17 & 6.69 & 7.60 & \textcolor{tablebest}{\textbf{11.58}} \\
TUM-19 & 9.80 & 9.99 & \textcolor{tablesecond}{\textbf{11.05}} & \textcolor{tablebest}{\textbf{12.00}} & 10.59 & 9.27 \\
TUM-20 & 10.13 & 10.21 & \textcolor{tablebest}{\textbf{14.74}} & 13.71 & 11.78 & \textcolor{tablesecond}{\textbf{14.15}} \\
TUM-22 & 12.20 & 13.29 & \textcolor{tablesecond}{\textbf{17.54}} & 17.35 & 11.40 & \textcolor{tablebest}{\textbf{17.77}} \\
TUM-23 & 11.99 & 11.64 & 15.57 & \textcolor{tablesecond}{\textbf{15.62}} & 12.15 & \textcolor{tablebest}{\textbf{17.19}} \\
TUM-25 & \textcolor{tablesecond}{\textbf{11.84}} & 11.67 & 11.35 & 11.17 & 10.85 & \textcolor{tablebest}{\textbf{12.32}} \\
TUM-26 & 10.85 & 10.49 & 12.66 & \textcolor{tablebest}{\textbf{12.93}} & 10.14 & \textcolor{tablesecond}{\textbf{12.90}} \\
TUM-27 & 12.04 & 12.82 & 14.53 & \textcolor{tablesecond}{\textbf{14.59}} & 13.56 & \textcolor{tablebest}{\textbf{14.67}} \\
TUM-29 & 13.54 & 12.42 & \textcolor{tablebest}{\textbf{16.19}} & 15.12 & 13.23 & \textcolor{tablesecond}{\textbf{15.75}} \\
TUM-30 & 17.84 & \textcolor{tablesecond}{\textbf{18.24}} & 16.61 & 15.37 & 12.88 & \textcolor{tablebest}{\textbf{18.78}} \\
TUM-33 & 16.96 & 15.84 & 15.69 & \textcolor{tablebest}{\textbf{17.83}} & 9.75 & \textcolor{tablesecond}{\textbf{17.79}} \\
TUM-34 & \textcolor{tablebest}{\textbf{20.17}} & 18.92 & 17.69 & 16.08 & 11.81 & \textcolor{tablesecond}{\textbf{19.28}} \\
TUM-35 & \textcolor{tablesecond}{\textbf{15.34}} & \textcolor{tablebest}{\textbf{15.87}} & 15.16 & 13.95 & 10.62 & 13.75 \\
TUM-36 & 11.09 & \textcolor{tablesecond}{\textbf{11.82}} & 9.31 & 10.78 & 7.65 & \textcolor{tablebest}{\textbf{12.03}} \\
TUM-38 & 11.06 & 14.49 & 14.58 & \textcolor{tablesecond}{\textbf{14.83}} & 13.26 & \textcolor{tablebest}{\textbf{15.35}} \\
TUM-41 & 17.26 & 17.73 & \textcolor{tablebest}{\textbf{21.48}} & 19.47 & 14.97 & \textcolor{tablesecond}{\textbf{20.22}} \\
TUM-42 & \textcolor{tablesecond}{\textbf{17.94}} & \textcolor{tablebest}{\textbf{18.29}} & 16.86 & 15.81 & 10.56 & 16.98 \\
TUM-43 & \textcolor{tablebest}{\textbf{18.61}} & 15.92 & 15.87 & 16.50 & 13.06 & \textcolor{tablesecond}{\textbf{17.53}} \\
TUM-44 & 10.55 & 12.50 & \textcolor{tablesecond}{\textbf{13.53}} & 12.59 & 9.55 & \textcolor{tablebest}{\textbf{14.07}} \\
TUM-47 & 11.11 & 12.43 & \textcolor{tablesecond}{\textbf{13.56}} & 12.38 & 8.28 & \textcolor{tablebest}{\textbf{14.15}} \\
\bottomrule
\end{tabular*}
\endgroup
\end{table}
\begin{table}[!htb]
\centering
\caption{\textbf{TUM RGB-D: SSIM per sequence.}}
\label{tab:supp_system_tum_rgbd_ssim}
\begingroup
\footnotesize
\setlength{\tabcolsep}{2.3pt}
\renewcommand{\arraystretch}{1.04}
\begin{tabular*}{\columnwidth}{@{\extracolsep{\fill}}lrrrrrr@{}}
\toprule
Sequence & \shortstack{Photo-\\SLAM} & CaRtGS & \shortstack{VIGS-\\SLAM} & \shortstack{HI-\\SLAM2} & \shortstack{Splat-\\SLAM} & Ours \\
\midrule
TUM-03 & 0.441 & \textcolor{tablesecond}{\textbf{0.465}} & 0.410 & 0.420 & 0.153 & \textcolor{tablebest}{\textbf{0.549}} \\
TUM-05 & \textcolor{tablesecond}{\textbf{0.430}} & 0.420 & 0.424 & 0.306 & 0.254 & \textcolor{tablebest}{\textbf{0.478}} \\
TUM-07 & \textcolor{tablebest}{\textbf{0.441}} & 0.372 & 0.313 & 0.154 & 0.017 & \textcolor{tablesecond}{\textbf{0.411}} \\
TUM-08 & 0.362 & 0.397 & \textcolor{tablesecond}{\textbf{0.427}} & 0.391 & 0.304 & \textcolor{tablebest}{\textbf{0.455}} \\
TUM-09 & 0.458 & 0.469 & 0.464 & \textcolor{tablesecond}{\textbf{0.499}} & 0.289 & \textcolor{tablebest}{\textbf{0.530}} \\
TUM-10 & \textcolor{tablebest}{\textbf{0.516}} & 0.431 & 0.379 & \textcolor{tablesecond}{\textbf{0.471}} & 0.279 & 0.343 \\
TUM-11 & \textcolor{tablebest}{\textbf{0.430}} & 0.365 & \textcolor{tablesecond}{\textbf{0.423}} & 0.178 & 0.238 & 0.422 \\
TUM-13 & 0.482 & 0.499 & \textcolor{tablesecond}{\textbf{0.504}} & 0.484 & 0.413 & \textcolor{tablebest}{\textbf{0.527}} \\
TUM-14 & 0.442 & 0.468 & 0.478 & \textcolor{tablesecond}{\textbf{0.479}} & 0.375 & \textcolor{tablebest}{\textbf{0.507}} \\
TUM-15 & 0.627 & 0.612 & \textcolor{tablesecond}{\textbf{0.672}} & 0.659 & 0.472 & \textcolor{tablebest}{\textbf{0.696}} \\
TUM-16 & 0.605 & 0.615 & 0.613 & \textcolor{tablesecond}{\textbf{0.630}} & 0.497 & \textcolor{tablebest}{\textbf{0.642}} \\
TUM-17 & 0.667 & 0.637 & 0.643 & \textcolor{tablesecond}{\textbf{0.687}} & 0.619 & \textcolor{tablebest}{\textbf{0.705}} \\
TUM-18 & \textcolor{tablebest}{\textbf{0.593}} & 0.542 & 0.488 & 0.142 & 0.187 & \textcolor{tablesecond}{\textbf{0.561}} \\
TUM-19 & 0.474 & \textcolor{tablesecond}{\textbf{0.483}} & \textcolor{tablebest}{\textbf{0.508}} & 0.426 & 0.369 & 0.433 \\
TUM-20 & \textcolor{tablesecond}{\textbf{0.585}} & 0.565 & 0.583 & 0.546 & 0.452 & \textcolor{tablebest}{\textbf{0.593}} \\
TUM-22 & 0.553 & 0.628 & \textcolor{tablesecond}{\textbf{0.668}} & 0.656 & 0.343 & \textcolor{tablebest}{\textbf{0.674}} \\
TUM-23 & 0.578 & 0.578 & 0.588 & \textcolor{tablesecond}{\textbf{0.615}} & 0.425 & \textcolor{tablebest}{\textbf{0.651}} \\
TUM-25 & \textcolor{tablesecond}{\textbf{0.556}} & \textcolor{tablebest}{\textbf{0.557}} & 0.508 & 0.434 & 0.376 & 0.531 \\
TUM-26 & \textcolor{tablebest}{\textbf{0.503}} & 0.437 & 0.477 & 0.481 & 0.380 & \textcolor{tablesecond}{\textbf{0.486}} \\
TUM-27 & 0.467 & 0.485 & 0.514 & \textcolor{tablesecond}{\textbf{0.519}} & 0.496 & \textcolor{tablebest}{\textbf{0.520}} \\
TUM-29 & \textcolor{tablebest}{\textbf{0.724}} & 0.639 & 0.664 & 0.673 & 0.656 & \textcolor{tablesecond}{\textbf{0.693}} \\
TUM-30 & 0.634 & \textcolor{tablesecond}{\textbf{0.643}} & 0.588 & 0.560 & 0.519 & \textcolor{tablebest}{\textbf{0.670}} \\
TUM-33 & \textcolor{tablesecond}{\textbf{0.672}} & 0.650 & 0.630 & \textcolor{tablebest}{\textbf{0.677}} & 0.389 & 0.671 \\
TUM-34 & \textcolor{tablesecond}{\textbf{0.681}} & \textcolor{tablebest}{\textbf{0.689}} & 0.634 & 0.546 & 0.425 & 0.666 \\
TUM-35 & \textcolor{tablesecond}{\textbf{0.539}} & \textcolor{tablebest}{\textbf{0.555}} & 0.507 & 0.461 & 0.336 & 0.466 \\
TUM-36 & \textcolor{tablebest}{\textbf{0.453}} & \textcolor{tablesecond}{\textbf{0.411}} & 0.318 & 0.339 & 0.142 & 0.409 \\
TUM-38 & 0.403 & 0.410 & 0.416 & \textcolor{tablesecond}{\textbf{0.420}} & 0.368 & \textcolor{tablebest}{\textbf{0.460}} \\
TUM-41 & 0.546 & 0.556 & \textcolor{tablebest}{\textbf{0.713}} & 0.618 & 0.450 & \textcolor{tablesecond}{\textbf{0.665}} \\
TUM-42 & \textcolor{tablesecond}{\textbf{0.641}} & \textcolor{tablebest}{\textbf{0.646}} & 0.595 & 0.574 & 0.366 & 0.613 \\
TUM-43 & \textcolor{tablebest}{\textbf{0.700}} & 0.654 & 0.630 & 0.644 & 0.581 & \textcolor{tablesecond}{\textbf{0.681}} \\
TUM-44 & 0.426 & \textcolor{tablesecond}{\textbf{0.453}} & 0.440 & 0.394 & 0.308 & \textcolor{tablebest}{\textbf{0.472}} \\
TUM-47 & \textcolor{tablesecond}{\textbf{0.398}} & 0.373 & 0.378 & 0.317 & 0.169 & \textcolor{tablebest}{\textbf{0.455}} \\
\bottomrule
\end{tabular*}
\endgroup
\end{table}
\begin{table}[!htb]
\centering
\caption{\textbf{TUM RGB-D: LPIPS per sequence.}}
\label{tab:supp_system_tum_rgbd_lpips}
\begingroup
\footnotesize
\setlength{\tabcolsep}{2.3pt}
\renewcommand{\arraystretch}{1.04}
\begin{tabular*}{\columnwidth}{@{\extracolsep{\fill}}lrrrrrr@{}}
\toprule
Sequence & \shortstack{Photo-\\SLAM} & CaRtGS & \shortstack{VIGS-\\SLAM} & \shortstack{HI-\\SLAM2} & \shortstack{Splat-\\SLAM} & Ours \\
\midrule
TUM-03 & 0.675 & \textcolor{tablesecond}{\textbf{0.610}} & 0.683 & 0.645 & 0.867 & \textcolor{tablebest}{\textbf{0.535}} \\
TUM-05 & \textcolor{tablesecond}{\textbf{0.658}} & 0.722 & 0.663 & 0.748 & 0.768 & \textcolor{tablebest}{\textbf{0.603}} \\
TUM-07 & 0.834 & \textcolor{tablesecond}{\textbf{0.798}} & 0.861 & 0.903 & 0.968 & \textcolor{tablebest}{\textbf{0.748}} \\
TUM-08 & 0.774 & 0.797 & \textcolor{tablesecond}{\textbf{0.647}} & \textcolor{tablesecond}{\textbf{0.647}} & 0.778 & \textcolor{tablebest}{\textbf{0.635}} \\
TUM-09 & 0.543 & 0.519 & \textcolor{tablesecond}{\textbf{0.460}} & 0.470 & 0.739 & \textcolor{tablebest}{\textbf{0.331}} \\
TUM-10 & 0.851 & 0.874 & 0.870 & \textcolor{tablebest}{\textbf{0.657}} & \textcolor{tablesecond}{\textbf{0.797}} & 0.898 \\
TUM-11 & \textcolor{tablesecond}{\textbf{0.844}} & 0.877 & 0.871 & 0.876 & \textcolor{tablebest}{\textbf{0.788}} & 0.868 \\
TUM-13 & 0.530 & 0.534 & \textcolor{tablesecond}{\textbf{0.515}} & 0.546 & 0.673 & \textcolor{tablebest}{\textbf{0.502}} \\
TUM-14 & 0.703 & 0.600 & \textcolor{tablesecond}{\textbf{0.543}} & 0.545 & 0.636 & \textcolor{tablebest}{\textbf{0.513}} \\
TUM-15 & 0.787 & 0.798 & \textcolor{tablesecond}{\textbf{0.472}} & 0.503 & 0.649 & \textcolor{tablebest}{\textbf{0.447}} \\
TUM-16 & 0.715 & 0.692 & 0.579 & \textcolor{tablebest}{\textbf{0.551}} & 0.648 & \textcolor{tablesecond}{\textbf{0.572}} \\
TUM-17 & 0.792 & 0.811 & 0.649 & \textcolor{tablebest}{\textbf{0.600}} & 0.722 & \textcolor{tablesecond}{\textbf{0.605}} \\
TUM-18 & \textcolor{tablesecond}{\textbf{0.799}} & 0.841 & 0.831 & 0.900 & 0.887 & \textcolor{tablebest}{\textbf{0.784}} \\
TUM-19 & 0.857 & 0.840 & \textcolor{tablebest}{\textbf{0.663}} & \textcolor{tablesecond}{\textbf{0.694}} & 0.718 & 0.843 \\
TUM-20 & 0.774 & 0.776 & \textcolor{tablebest}{\textbf{0.607}} & 0.636 & 0.698 & \textcolor{tablesecond}{\textbf{0.633}} \\
TUM-22 & 0.801 & 0.792 & \textcolor{tablesecond}{\textbf{0.565}} & 0.617 & 0.721 & \textcolor{tablebest}{\textbf{0.482}} \\
TUM-23 & 0.807 & 0.826 & \textcolor{tablesecond}{\textbf{0.696}} & 0.700 & 0.739 & \textcolor{tablebest}{\textbf{0.584}} \\
TUM-25 & 0.824 & 0.831 & 0.824 & 0.836 & \textcolor{tablebest}{\textbf{0.785}} & \textcolor{tablesecond}{\textbf{0.800}} \\
TUM-26 & 0.858 & 0.717 & \textcolor{tablesecond}{\textbf{0.506}} & \textcolor{tablebest}{\textbf{0.500}} & 0.666 & 0.554 \\
TUM-27 & 0.578 & 0.459 & \textcolor{tablesecond}{\textbf{0.381}} & 0.393 & 0.428 & \textcolor{tablebest}{\textbf{0.369}} \\
TUM-29 & 0.694 & 0.755 & 0.649 & \textcolor{tablesecond}{\textbf{0.643}} & 0.680 & \textcolor{tablebest}{\textbf{0.564}} \\
TUM-30 & \textcolor{tablebest}{\textbf{0.330}} & \textcolor{tablesecond}{\textbf{0.338}} & 0.455 & 0.475 & 0.695 & 0.372 \\
TUM-33 & \textcolor{tablesecond}{\textbf{0.451}} & 0.510 & 0.588 & \textcolor{tablebest}{\textbf{0.430}} & 0.712 & 0.465 \\
TUM-34 & \textcolor{tablebest}{\textbf{0.264}} & 0.368 & 0.497 & 0.498 & 0.660 & \textcolor{tablesecond}{\textbf{0.358}} \\
TUM-35 & \textcolor{tablesecond}{\textbf{0.418}} & \textcolor{tablebest}{\textbf{0.414}} & 0.483 & 0.501 & 0.685 & 0.535 \\
TUM-36 & 0.886 & \textcolor{tablesecond}{\textbf{0.685}} & 0.742 & 0.702 & 0.849 & \textcolor{tablebest}{\textbf{0.680}} \\
TUM-38 & 0.891 & \textcolor{tablesecond}{\textbf{0.385}} & 0.427 & \textcolor{tablebest}{\textbf{0.354}} & 0.454 & 0.435 \\
TUM-41 & 0.261 & 0.232 & \textcolor{tablebest}{\textbf{0.212}} & 0.226 & 0.466 & \textcolor{tablesecond}{\textbf{0.224}} \\
TUM-42 & \textcolor{tablesecond}{\textbf{0.330}} & \textcolor{tablebest}{\textbf{0.308}} & 0.435 & 0.496 & 0.790 & 0.395 \\
TUM-43 & \textcolor{tablebest}{\textbf{0.387}} & 0.553 & 0.586 & 0.529 & 0.730 & \textcolor{tablesecond}{\textbf{0.483}} \\
TUM-44 & 0.909 & 0.592 & 0.569 & \textcolor{tablesecond}{\textbf{0.555}} & 0.802 & \textcolor{tablebest}{\textbf{0.522}} \\
TUM-47 & 0.916 & 0.569 & \textcolor{tablesecond}{\textbf{0.523}} & 0.537 & 0.825 & \textcolor{tablebest}{\textbf{0.493}} \\
\bottomrule
\end{tabular*}
\endgroup
\end{table}
\begin{table}[!htb]
\centering
\caption{\textbf{TUM RGB-D: depth MAE (m) per sequence.}}
\label{tab:supp_system_tum_rgbd_mae_m}
\begingroup
\footnotesize
\setlength{\tabcolsep}{2.3pt}
\renewcommand{\arraystretch}{1.04}
\begin{tabular*}{\columnwidth}{@{\extracolsep{\fill}}lrrrrrr@{}}
\toprule
Sequence & \shortstack{Photo-\\SLAM} & CaRtGS & \shortstack{VIGS-\\SLAM} & \shortstack{HI-\\SLAM2} & \shortstack{Splat-\\SLAM} & Ours \\
\midrule
TUM-03 & 0.633 & 0.523 & 0.264 & \textcolor{tablesecond}{\textbf{0.206}} & 0.551 & \textcolor{tablebest}{\textbf{0.156}} \\
TUM-05 & 0.629 & 0.760 & \textcolor{tablebest}{\textbf{0.354}} & 0.636 & 0.397 & \textcolor{tablesecond}{\textbf{0.360}} \\
TUM-07 & \textcolor{tablesecond}{\textbf{1.141}} & 1.225 & 1.842 & 1.738 & 2.682 & \textcolor{tablebest}{\textbf{0.566}} \\
TUM-08 & 1.517 & 1.451 & \textcolor{tablesecond}{\textbf{0.301}} & \textcolor{tablebest}{\textbf{0.287}} & 0.559 & 0.308 \\
TUM-09 & 0.190 & 0.205 & \textcolor{tablesecond}{\textbf{0.128}} & 0.178 & 0.357 & \textcolor{tablebest}{\textbf{0.095}} \\
TUM-19 & 4.787 & 4.085 & \textcolor{tablebest}{\textbf{0.710}} & \textcolor{tablesecond}{\textbf{0.723}} & 1.142 & 3.337 \\
TUM-20 & 2.443 & 2.236 & \textcolor{tablebest}{\textbf{0.484}} & 0.644 & \textcolor{tablesecond}{\textbf{0.560}} & 1.060 \\
TUM-29 & 2.973 & 13.460 & \textcolor{tablesecond}{\textbf{0.691}} & 1.113 & 1.878 & \textcolor{tablebest}{\textbf{0.539}} \\
TUM-30 & 0.507 & 0.327 & \textcolor{tablesecond}{\textbf{0.159}} & 0.224 & 0.603 & \textcolor{tablebest}{\textbf{0.141}} \\
TUM-33 & 1.368 & 0.674 & \textcolor{tablesecond}{\textbf{0.080}} & 0.115 & 0.263 & \textcolor{tablebest}{\textbf{0.063}} \\
TUM-34 & 0.352 & 0.546 & \textcolor{tablesecond}{\textbf{0.018}} & 0.027 & 0.100 & \textcolor{tablebest}{\textbf{0.016}} \\
TUM-35 & 0.515 & 0.623 & \textcolor{tablebest}{\textbf{0.346}} & 0.367 & 0.528 & \textcolor{tablesecond}{\textbf{0.359}} \\
TUM-36 & 2.604 & 1.093 & 1.296 & \textcolor{tablebest}{\textbf{0.803}} & 1.438 & \textcolor{tablesecond}{\textbf{0.946}} \\
TUM-38 & 2.595 & 0.945 & 0.340 & \textcolor{tablebest}{\textbf{0.270}} & 0.474 & \textcolor{tablesecond}{\textbf{0.333}} \\
TUM-41 & 0.282 & 0.223 & \textcolor{tablesecond}{\textbf{0.055}} & 0.070 & 0.111 & \textcolor{tablebest}{\textbf{0.051}} \\
TUM-42 & 0.349 & 0.303 & \textcolor{tablesecond}{\textbf{0.118}} & 0.146 & 0.306 & \textcolor{tablebest}{\textbf{0.111}} \\
TUM-43 & \textcolor{tablesecond}{\textbf{0.904}} & 1.218 & 1.175 & 1.075 & 1.707 & \textcolor{tablebest}{\textbf{0.722}} \\
TUM-44 & 4.558 & 1.296 & 0.829 & \textcolor{tablesecond}{\textbf{0.751}} & 1.560 & \textcolor{tablebest}{\textbf{0.697}} \\
TUM-47 & 2.746 & 1.519 & \textcolor{tablesecond}{\textbf{1.069}} & 1.086 & 1.613 & \textcolor{tablebest}{\textbf{1.016}} \\
\bottomrule
\end{tabular*}
\endgroup
\end{table}
\begin{table}[!htb]
\centering
\caption{\textbf{TUM RGB-D: depth coverage (\%) per sequence.}}
\label{tab:supp_system_tum_rgbd_coverage}
\begingroup
\footnotesize
\setlength{\tabcolsep}{2.3pt}
\renewcommand{\arraystretch}{1.04}
\begin{tabular*}{\columnwidth}{@{\extracolsep{\fill}}lrrrrrr@{}}
\toprule
Sequence & \shortstack{Photo-\\SLAM} & CaRtGS & \shortstack{VIGS-\\SLAM} & \shortstack{HI-\\SLAM2} & \shortstack{Splat-\\SLAM} & Ours \\
\midrule
TUM-03 & \textcolor{tablebest}{\textbf{100.00}} & 98.92 & 92.32 & 91.05 & 50.36 & \textcolor{tablesecond}{\textbf{99.83}} \\
TUM-05 & \textcolor{tablebest}{\textbf{100.00}} & \textcolor{tablesecond}{\textbf{99.82}} & 83.23 & 83.15 & 60.36 & 99.03 \\
TUM-07 & \textcolor{tablebest}{\textbf{99.75}} & 93.91 & 51.07 & 38.21 & 0.09 & \textcolor{tablesecond}{\textbf{95.57}} \\
TUM-08 & \textcolor{tablesecond}{\textbf{99.74}} & 96.30 & 85.46 & 86.33 & 71.87 & \textcolor{tablebest}{\textbf{99.76}} \\
TUM-09 & \textcolor{tablebest}{\textbf{100.00}} & \textcolor{tablesecond}{\textbf{99.90}} & 94.25 & 95.63 & 93.04 & 99.29 \\
TUM-19 & \textcolor{tablesecond}{\textbf{92.47}} & 90.73 & 77.54 & 74.34 & 78.54 & \textcolor{tablebest}{\textbf{97.87}} \\
TUM-20 & \textcolor{tablebest}{\textbf{100.00}} & \textcolor{tablesecond}{\textbf{99.99}} & 99.17 & 92.60 & 73.41 & 94.19 \\
TUM-29 & \textcolor{tablebest}{\textbf{99.99}} & 84.51 & 99.49 & 99.14 & 98.45 & \textcolor{tablesecond}{\textbf{99.83}} \\
TUM-30 & \textcolor{tablebest}{\textbf{100.00}} & \textcolor{tablebest}{\textbf{100.00}} & 99.04 & 99.28 & 97.79 & \textcolor{tablesecond}{\textbf{99.89}} \\
TUM-33 & \textcolor{tablebest}{\textbf{100.00}} & 98.54 & 97.74 & 98.09 & 61.02 & \textcolor{tablesecond}{\textbf{98.92}} \\
TUM-34 & \textcolor{tablebest}{\textbf{100.00}} & \textcolor{tablebest}{\textbf{100.00}} & 94.93 & 86.90 & 60.34 & \textcolor{tablesecond}{\textbf{97.67}} \\
TUM-35 & \textcolor{tablebest}{\textbf{100.00}} & \textcolor{tablesecond}{\textbf{99.96}} & 99.71 & 98.93 & 93.89 & 99.90 \\
TUM-36 & \textcolor{tablebest}{\textbf{100.00}} & 90.77 & 88.64 & 91.33 & 39.04 & \textcolor{tablesecond}{\textbf{99.27}} \\
TUM-38 & \textcolor{tablebest}{\textbf{100.00}} & \textcolor{tablebest}{\textbf{100.00}} & 97.55 & \textcolor{tablesecond}{\textbf{99.76}} & 98.39 & 99.38 \\
TUM-41 & \textcolor{tablebest}{\textbf{100.00}} & \textcolor{tablebest}{\textbf{100.00}} & \textcolor{tablesecond}{\textbf{99.88}} & 99.80 & 94.17 & 99.72 \\
TUM-42 & \textcolor{tablebest}{\textbf{100.00}} & \textcolor{tablesecond}{\textbf{99.99}} & 98.78 & 98.45 & 65.87 & 99.85 \\
TUM-43 & \textcolor{tablebest}{\textbf{99.99}} & \textcolor{tablesecond}{\textbf{99.87}} & 98.67 & 98.22 & 95.46 & 98.74 \\
TUM-44 & 88.58 & \textcolor{tablebest}{\textbf{99.93}} & 97.61 & 98.05 & 94.07 & \textcolor{tablesecond}{\textbf{98.97}} \\
TUM-47 & \textcolor{tablebest}{\textbf{100.00}} & \textcolor{tablebest}{\textbf{100.00}} & \textcolor{tablesecond}{\textbf{99.89}} & 99.70 & 54.95 & 99.77 \\
\bottomrule
\end{tabular*}
\endgroup
\end{table}
\begin{table}[!htb]
\centering
\caption{\textbf{TUM RGB-D: within 5\,cm (\%) per sequence.}}
\label{tab:supp_system_tum_rgbd_within5cm}
\begingroup
\footnotesize
\setlength{\tabcolsep}{2.3pt}
\renewcommand{\arraystretch}{1.04}
\begin{tabular*}{\columnwidth}{@{\extracolsep{\fill}}lrrrrrr@{}}
\toprule
Sequence & \shortstack{Photo-\\SLAM} & CaRtGS & \shortstack{VIGS-\\SLAM} & \shortstack{HI-\\SLAM2} & \shortstack{Splat-\\SLAM} & Ours \\
\midrule
TUM-03 & 21.64 & 24.65 & 32.07 & \textcolor{tablesecond}{\textbf{37.05}} & 3.62 & \textcolor{tablebest}{\textbf{54.83}} \\
TUM-05 & 11.95 & 10.24 & \textcolor{tablesecond}{\textbf{15.79}} & 8.31 & 11.33 & \textcolor{tablebest}{\textbf{21.91}} \\
TUM-07 & 1.20 & \textcolor{tablesecond}{\textbf{3.62}} & 0.46 & 0.07 & 0.00 & \textcolor{tablebest}{\textbf{4.43}} \\
TUM-08 & 1.34 & 1.48 & \textcolor{tablesecond}{\textbf{20.77}} & \textcolor{tablebest}{\textbf{22.75}} & 8.18 & 19.26 \\
TUM-09 & 36.49 & 35.67 & \textcolor{tablesecond}{\textbf{58.08}} & 51.94 & 14.83 & \textcolor{tablebest}{\textbf{61.61}} \\
TUM-19 & 0.11 & 0.36 & \textcolor{tablesecond}{\textbf{6.95}} & \textcolor{tablebest}{\textbf{7.10}} & 3.01 & 1.10 \\
TUM-20 & 1.17 & 0.53 & \textcolor{tablebest}{\textbf{36.27}} & \textcolor{tablesecond}{\textbf{28.03}} & 22.43 & 26.59 \\
TUM-29 & 0.10 & 0.48 & \textcolor{tablesecond}{\textbf{11.24}} & 6.38 & 7.30 & \textcolor{tablebest}{\textbf{17.01}} \\
TUM-30 & 40.98 & 30.49 & \textcolor{tablesecond}{\textbf{55.06}} & 47.19 & 25.59 & \textcolor{tablebest}{\textbf{64.28}} \\
TUM-33 & 13.73 & 10.50 & \textcolor{tablesecond}{\textbf{50.84}} & 43.84 & 7.69 & \textcolor{tablebest}{\textbf{58.68}} \\
TUM-34 & 44.70 & 30.20 & \textcolor{tablesecond}{\textbf{89.81}} & 75.80 & 37.88 & \textcolor{tablebest}{\textbf{94.54}} \\
TUM-35 & 19.81 & 11.56 & \textcolor{tablebest}{\textbf{27.17}} & \textcolor{tablesecond}{\textbf{20.50}} & 19.87 & 19.61 \\
TUM-36 & 0.00 & \textcolor{tablesecond}{\textbf{6.93}} & 1.15 & \textcolor{tablebest}{\textbf{10.63}} & 2.57 & 4.90 \\
TUM-38 & 0.00 & 10.83 & \textcolor{tablesecond}{\textbf{34.64}} & \textcolor{tablebest}{\textbf{35.89}} & 21.18 & 27.75 \\
TUM-41 & 28.59 & 27.38 & \textcolor{tablebest}{\textbf{85.09}} & 79.77 & 52.58 & \textcolor{tablesecond}{\textbf{82.59}} \\
TUM-42 & 31.46 & 24.26 & \textcolor{tablesecond}{\textbf{75.24}} & 68.00 & 16.96 & \textcolor{tablebest}{\textbf{77.68}} \\
TUM-43 & \textcolor{tablesecond}{\textbf{30.63}} & 15.21 & 26.24 & 29.00 & 15.75 & \textcolor{tablebest}{\textbf{33.54}} \\
TUM-44 & 0.03 & 3.00 & 5.17 & \textcolor{tablesecond}{\textbf{14.84}} & 3.07 & \textcolor{tablebest}{\textbf{19.93}} \\
TUM-47 & 0.00 & 5.28 & 9.76 & \textcolor{tablesecond}{\textbf{18.19}} & 1.63 & \textcolor{tablebest}{\textbf{22.81}} \\
\bottomrule
\end{tabular*}
\endgroup
\end{table}

\FloatBarrier
\subsection{FAST-LIVO2}
\label{sec:supp_tables_fast_livo2}
\begin{table}[!htb]
\centering
\caption{\textbf{FAST-LIVO2: PSNR per sequence.}}
\label{tab:supp_system_fast_livo2_psnr}
\begingroup
\footnotesize
\setlength{\tabcolsep}{2.3pt}
\renewcommand{\arraystretch}{1.04}
\begin{tabular*}{\columnwidth}{@{\extracolsep{\fill}}lrrrrrr@{}}
\toprule
Sequence & \shortstack{Photo-\\SLAM} & CaRtGS & \shortstack{VIGS-\\SLAM} & \shortstack{HI-\\SLAM2} & \shortstack{Splat-\\SLAM} & Ours \\
\midrule
FAST-01 & 11.44 & 14.13 & 16.84 & \textcolor{tablebest}{\textbf{19.64}} & 7.45 & \textcolor{tablesecond}{\textbf{18.52}} \\
FAST-02 & 10.43 & 9.77 & 15.66 & \textcolor{tablebest}{\textbf{17.76}} & 12.04 & \textcolor{tablesecond}{\textbf{17.19}} \\
FAST-03 & 17.60 & 17.92 & 17.92 & \textcolor{tablebest}{\textbf{22.09}} & 18.27 & \textcolor{tablesecond}{\textbf{22.08}} \\
FAST-04 & 12.08 & 11.83 & \textcolor{tablesecond}{\textbf{15.21}} & 13.17 & 10.49 & \textcolor{tablebest}{\textbf{17.17}} \\
FAST-05 & 12.00 & 13.05 & 16.87 & \textcolor{tablesecond}{\textbf{17.73}} & 11.79 & \textcolor{tablebest}{\textbf{19.69}} \\
\bottomrule
\end{tabular*}
\endgroup
\end{table}
\begin{table}[!htb]
\centering
\caption{\textbf{FAST-LIVO2: SSIM per sequence.}}
\label{tab:supp_system_fast_livo2_ssim}
\begingroup
\footnotesize
\setlength{\tabcolsep}{2.3pt}
\renewcommand{\arraystretch}{1.04}
\begin{tabular*}{\columnwidth}{@{\extracolsep{\fill}}lrrrrrr@{}}
\toprule
Sequence & \shortstack{Photo-\\SLAM} & CaRtGS & \shortstack{VIGS-\\SLAM} & \shortstack{HI-\\SLAM2} & \shortstack{Splat-\\SLAM} & Ours \\
\midrule
FAST-01 & 0.476 & 0.524 & 0.524 & \textcolor{tablebest}{\textbf{0.658}} & 0.259 & \textcolor{tablesecond}{\textbf{0.592}} \\
FAST-02 & 0.378 & 0.362 & \textcolor{tablesecond}{\textbf{0.522}} & \textcolor{tablebest}{\textbf{0.524}} & 0.318 & 0.482 \\
FAST-03 & 0.471 & 0.481 & 0.462 & \textcolor{tablebest}{\textbf{0.547}} & 0.440 & \textcolor{tablesecond}{\textbf{0.523}} \\
FAST-04 & \textcolor{tablesecond}{\textbf{0.422}} & 0.420 & 0.398 & 0.337 & 0.235 & \textcolor{tablebest}{\textbf{0.441}} \\
FAST-05 & 0.411 & 0.400 & 0.476 & \textcolor{tablesecond}{\textbf{0.511}} & 0.294 & \textcolor{tablebest}{\textbf{0.574}} \\
\bottomrule
\end{tabular*}
\endgroup
\end{table}
\begin{table}[!htb]
\centering
\caption{\textbf{FAST-LIVO2: LPIPS per sequence.}}
\label{tab:supp_system_fast_livo2_lpips}
\begingroup
\footnotesize
\setlength{\tabcolsep}{2.3pt}
\renewcommand{\arraystretch}{1.04}
\begin{tabular*}{\columnwidth}{@{\extracolsep{\fill}}lrrrrrr@{}}
\toprule
Sequence & \shortstack{Photo-\\SLAM} & CaRtGS & \shortstack{VIGS-\\SLAM} & \shortstack{HI-\\SLAM2} & \shortstack{Splat-\\SLAM} & Ours \\
\midrule
FAST-01 & 0.710 & 0.617 & 0.447 & \textcolor{tablesecond}{\textbf{0.354}} & 0.848 & \textcolor{tablebest}{\textbf{0.247}} \\
FAST-02 & 0.866 & 0.838 & 0.513 & \textcolor{tablesecond}{\textbf{0.424}} & 0.733 & \textcolor{tablebest}{\textbf{0.290}} \\
FAST-03 & 0.651 & 0.637 & 0.445 & \textcolor{tablesecond}{\textbf{0.372}} & 0.451 & \textcolor{tablebest}{\textbf{0.319}} \\
FAST-04 & 0.854 & 0.865 & \textcolor{tablesecond}{\textbf{0.349}} & 0.631 & 0.672 & \textcolor{tablebest}{\textbf{0.236}} \\
FAST-05 & 0.815 & 0.719 & \textcolor{tablesecond}{\textbf{0.463}} & 0.498 & 0.648 & \textcolor{tablebest}{\textbf{0.313}} \\
\bottomrule
\end{tabular*}
\endgroup
\end{table}
\begin{table}[!htb]
\centering
\caption{\textbf{FAST-LIVO2: depth MAE (m) per sequence.}}
\label{tab:supp_system_fast_livo2_mae_m}
\begingroup
\footnotesize
\setlength{\tabcolsep}{2.3pt}
\renewcommand{\arraystretch}{1.04}
\begin{tabular*}{\columnwidth}{@{\extracolsep{\fill}}lrrrrrr@{}}
\toprule
Sequence & \shortstack{Photo-\\SLAM} & CaRtGS & \shortstack{VIGS-\\SLAM} & \shortstack{HI-\\SLAM2} & \shortstack{Splat-\\SLAM} & Ours \\
\midrule
FAST-01 & 7.636 & 5.014 & 1.259 & \textcolor{tablesecond}{\textbf{1.000}} & 6.900 & \textcolor{tablebest}{\textbf{0.718}} \\
FAST-02 & 19.964 & 7.598 & 0.790 & \textcolor{tablesecond}{\textbf{0.495}} & 11.085 & \textcolor{tablebest}{\textbf{0.486}} \\
FAST-03 & 3.687 & 3.354 & 0.565 & \textcolor{tablebest}{\textbf{0.535}} & 0.758 & \textcolor{tablesecond}{\textbf{0.539}} \\
FAST-04 & 6.796 & 33.190 & \textcolor{tablesecond}{\textbf{0.496}} & 0.700 & 0.916 & \textcolor{tablebest}{\textbf{0.433}} \\
FAST-05 & 13.381 & 9.995 & 1.202 & \textcolor{tablesecond}{\textbf{1.074}} & 1.709 & \textcolor{tablebest}{\textbf{0.615}} \\
\bottomrule
\end{tabular*}
\endgroup
\end{table}
\begin{table}[!htb]
\centering
\caption{\textbf{FAST-LIVO2: depth coverage (\%) per sequence.}}
\label{tab:supp_system_fast_livo2_coverage}
\begingroup
\footnotesize
\setlength{\tabcolsep}{2.3pt}
\renewcommand{\arraystretch}{1.04}
\begin{tabular*}{\columnwidth}{@{\extracolsep{\fill}}lrrrrrr@{}}
\toprule
Sequence & \shortstack{Photo-\\SLAM} & CaRtGS & \shortstack{VIGS-\\SLAM} & \shortstack{HI-\\SLAM2} & \shortstack{Splat-\\SLAM} & Ours \\
\midrule
FAST-01 & \textcolor{tablebest}{\textbf{100.00}} & 99.81 & 99.07 & \textcolor{tablesecond}{\textbf{99.92}} & 84.93 & 99.88 \\
FAST-02 & 90.31 & 95.39 & 68.96 & \textcolor{tablesecond}{\textbf{96.74}} & 57.94 & \textcolor{tablebest}{\textbf{99.38}} \\
FAST-03 & \textcolor{tablebest}{\textbf{99.71}} & \textcolor{tablesecond}{\textbf{99.16}} & 92.12 & 96.37 & 91.29 & 98.43 \\
FAST-04 & 95.96 & 95.57 & 97.53 & \textcolor{tablesecond}{\textbf{98.72}} & 53.57 & \textcolor{tablebest}{\textbf{99.69}} \\
FAST-05 & 99.54 & \textcolor{tablesecond}{\textbf{99.88}} & 94.66 & 99.63 & 75.95 & \textcolor{tablebest}{\textbf{99.89}} \\
\bottomrule
\end{tabular*}
\endgroup
\end{table}
\begin{table}[!htb]
\centering
\caption{\textbf{FAST-LIVO2: within 5\,cm (\%) per sequence.}}
\label{tab:supp_system_fast_livo2_within5cm}
\begingroup
\footnotesize
\setlength{\tabcolsep}{2.3pt}
\renewcommand{\arraystretch}{1.04}
\begin{tabular*}{\columnwidth}{@{\extracolsep{\fill}}lrrrrrr@{}}
\toprule
Sequence & \shortstack{Photo-\\SLAM} & CaRtGS & \shortstack{VIGS-\\SLAM} & \shortstack{HI-\\SLAM2} & \shortstack{Splat-\\SLAM} & Ours \\
\midrule
FAST-01 & 0.93 & 0.73 & 14.93 & \textcolor{tablesecond}{\textbf{21.94}} & 0.46 & \textcolor{tablebest}{\textbf{29.61}} \\
FAST-02 & 0.26 & 0.55 & 27.61 & \textcolor{tablebest}{\textbf{42.55}} & 3.44 & \textcolor{tablesecond}{\textbf{29.44}} \\
FAST-03 & 1.57 & 1.06 & \textcolor{tablebest}{\textbf{30.66}} & \textcolor{tablesecond}{\textbf{26.02}} & 21.42 & 22.39 \\
FAST-04 & 0.43 & 0.50 & \textcolor{tablebest}{\textbf{21.62}} & 13.58 & 6.65 & \textcolor{tablesecond}{\textbf{17.27}} \\
FAST-05 & 0.10 & 0.20 & 4.47 & \textcolor{tablesecond}{\textbf{7.04}} & 2.61 & \textcolor{tablebest}{\textbf{16.48}} \\
\bottomrule
\end{tabular*}
\endgroup
\end{table}
\FloatBarrier

\end{document}